\documentclass[fleqn,10pt]{wlscirep}
\usepackage[utf8]{inputenc}
\usepackage[T1]{fontenc}
\usepackage{bm}
\usepackage{multirow} 

\title{Factorization Machine with Quadratic-Optimization Annealing for RNA Inverse Folding and Evaluation of Binary-Integer Encoding and Nucleotide Assignment}

\author[1,2*]{Shuta Kikuchi}
\author[1,2,3,4]{Shu Tanaka}
\affil[1]{Graduate School of Science and Technology, Keio University, Yokohama, Kanagawa 223-8522, Japan}
\affil[2]{Keio University Sustainable Quantum Artificial Intelligence Center (KSQAIC), Keio University, Minato-ku, Tokyo 108-8345, Japan}
\affil[3]{Department of Applied Physics and Physico-Informatics, Keio University, Yokohama, Kanagawa 223-8522, Japan}
\affil[4]{Human Biology-Microbiome-Quantum Research Center (WPI-Bio2Q), Keio University, Minato-ku, Tokyo 108-8345, Japan}

\affil[*]{kikuchi.shuta@keio.jp}

\begin{abstract}
The RNA inverse folding problem aims to identify nucleotide sequences that preferentially adopt a given target secondary structure. 
While various heuristic and machine learning-based approaches have been proposed, many require a large number of sequence evaluations, which limits their applicability when experimental validation is costly. 
We propose a method to solve the problem using a factorization machine with quadratic-optimization annealing (FMQA).
FMQA is a discrete black-box optimization method reported to obtain high-quality solutions with a limited number of evaluations.
Applying FMQA to the problem requires converting nucleotides into binary variables. 
However, the influence of integer-to-nucleotide assignments and binary-integer encoding on the performance of FMQA has not been thoroughly investigated, even though such choices determine the structure of the surrogate model and the search landscape, and thus can directly affect solution quality.
Therefore, this study aims both to establish a novel FMQA framework for RNA inverse folding and to analyze the effects of these assignments and encoding methods.
We evaluated all 24 possible assignments of the four nucleotides to the ordered integers (0-3), in combination with four binary-integer encoding methods.
Our results demonstrated that one-hot and domain-wall encodings outperform binary and unary encodings in terms of the normalized ensemble defect value.
In domain-wall encoding, nucleotides assigned to the boundary integers (0 and 3) appeared with higher frequency.
In the RNA inverse folding problem, assigning guanine and cytosine to these boundary integers promoted their enrichment in stem regions, which led to more thermodynamically stable secondary structures than those obtained with one-hot encoding.
\end{abstract}
\begin{document}

\flushbottom
\maketitle

\section*{Introduction}

Ribonucleic acid (RNA) is an essential molecule that plays an important role in fundamental biological processes, including transcription and translation~\cite{crick1970central}, cellular differentiation and development~\cite{morris2014rise}, catalyzing reactions~\cite{doudna2002chemical}, and controlling gene expression~\cite{serganov2007ribozymes}.
In addition, synthetic RNAs are currently utilized in various fields, such as mRNA vaccines~\cite{pardi2018mrna}, RNA aptamers~\cite{hamada2018silico}, genome editing~\cite{singh2017exploring}, biosensing~\cite{jaffrey2018rna}, ribozymes~\cite{bauer2006engineered}, and riboswitches~\cite{dixon2010reengineering}.

\begin{figure}[t]
    \centering
    \includegraphics[width=0.7\linewidth]{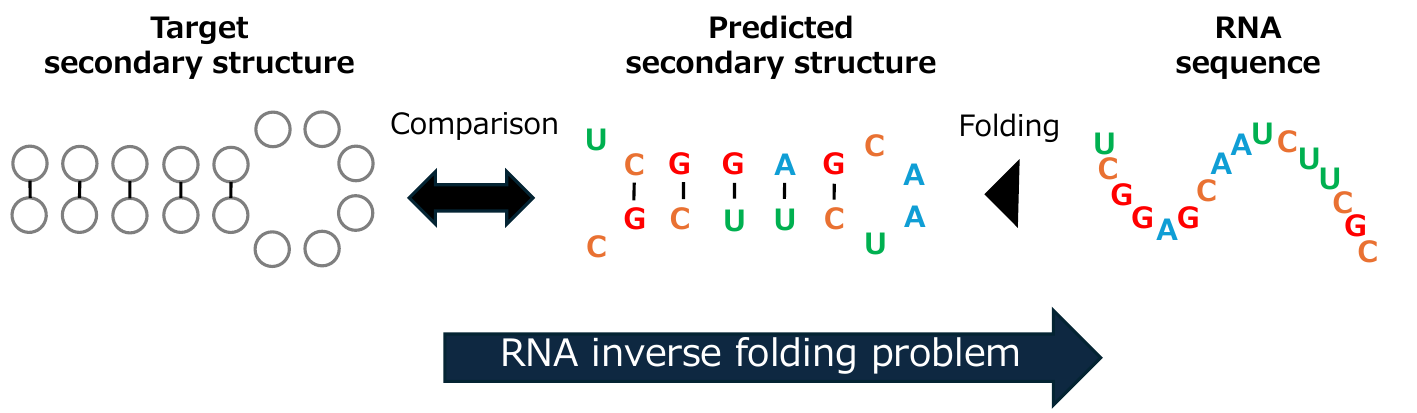}
    \caption{Overview of the RNA inverse folding problem. White circles represent arbitrary nucleotides in target secondary structure, and black lines indicate base pairs formed through hydrogen bonding. A, U, G, and C denote adenine, uracil, guanine, and cytosine, respectively.}
    \label{fig:RNA_inverse_folding}
\end{figure}

RNA function depends on its structure.
The primary structure of RNA refers to its linear sequence of four nucleotides linked by phosphodiester bonds.
The nucleotides consist of a $5$-carbon sugar ribose, a phosphate group, and one of four bases: adenine (A), uracil (U), guanine (G), and cytosine (C).
Under physiological conditions, nucleotides in the sequence interact through specific hydrogen bonds to form canonical Watson--Crick base pairs (A-U, G-C)~\cite{seeman1976rna, rosenberg1976rna} and, less commonly, wobble base pairs (U-G)~\cite{varani2000g}.
Those base-pairing interactions lead to the creation of the secondary structure.
Then, this secondary structure guides the formation of the three-dimensional structure, referred to as the tertiary structure.
Although RNA function ultimately depends on its three-dimensional structure, which is determined by the underlying secondary structure arising from base-pairing interactions, synthetic RNAs can be readily constructed with arbitrary nucleotide sequences~\cite{reese2005oligo}.
Consequently, the task of identifying RNA sequences that fold into a desired secondary structure (target structure) is known as the RNA inverse folding problem~\cite{hofacker1994fast}(Fig.~\ref{fig:RNA_inverse_folding}).
This problem is typically formulated as the search for an RNA sequence whose minimum free energy (MFE) structure matches the target structure.
Theoretical results have shown that the problem is NP-hard, both in its most general formulation and even under simplified energy models~\cite{schnall2008inverting, bonnet2020designing}.

Several approaches have been developed to solve the RNA inverse folding problem.
The pioneering tool, RNAinverse~\cite{hofacker1994fast}, uses an adaptive random walk to minimize the base pair distance between the MFE secondary structure of the current sequence and the target structure.
Other approaches use stochastic local search~\cite{andronescu2004new, busch2006info, zadeh2011nucleic, zadeh2011nupack}, genetic algorithms~\cite{taneda2010modena, lyngso2012frnakenstein, rubio2018multiobjective}, constraint programming~\cite{garcia2013rnaifold}, ant colony optimization~\cite{kleinkauf2015antarna}, and Monte Carlo methods~\cite{yang2017rna}.
In addition, deep learning-based approaches have been proposed, including supervised learning from human solutions~\cite{shi2018sentrna, koodli2019eternabrain}, reinforcement learning~\cite{eastman2018solving, runge2018learning}, and generative models~\cite{sumi2024deep}, and transformer-based deep representation learning~\cite{akiyama2022informative}.
These methods have advanced the field of the RNA inverse folding problem by achieving high success rates and producing high-quality sequences on difficult benchmark problems. 
However, the need for a large number of samples during the search process or for large-scale training data remains a limitation. 
Although digitally folded RNA sequences are commonly used for evaluation, wet-lab experiments benefit from approaches that limit the number of evaluations, given the considerable time and resource costs associated with experimental validation.

Black-box (BB) optimization methods that aim to reduce the number of expensive evaluations by combining surrogate modeling with iterative optimization have been actively studied~\cite{jones1998efficient, forrester2009recent, ramu2022survey}. 
These approaches construct a surrogate model from a limited number of observed samples and exploit it to efficiently search for better candidates, which leads to reduced overall evaluation cost.
Recently, factorization machine with quadratic-optimization annealing (FMQA)~\cite{tamura2025black} has been proposed as a discrete black-box optimization method that employs a factorization machine (FM)~\cite{rendle2010factorization} as a surrogate model and performs optimization using quadratic optimization solvers.
Originally, this framework is referred to as factorization machine with quantum annealing (FMQA)~\cite{kitai2020designing}.
Ising machines are commonly used as quadratic optimization solvers. 
Ising machines have attracted attention as accurate and efficient solvers for combinatorial optimization problems~\cite{mohseni2022ising}.
Various types of Ising machines have been developed.
Quantum annealing machines~\cite{Johnson2011} are implemented using superconducting circuits, and their internal algorithm is quantum annealing~\cite{kadowaki1998quantum, farhi2000quantum}.
Digital Ising machines~\cite{tsukamoto2017accelerator, okuyama2017ising, goto2021high, FixAE} are implemented using digital circuits, such as graphics processing units (GPU), field-programmable gate arrays (FPGA).
Their internal algorithms are based on simulated annealing (SA)~\cite{kirkpatrick1983optimization, johnson1991optimization}, simulated quantum annealing~\cite{das2005quantum}, and simulated bifurcations~\cite{goto2019combinatorial}.
To tackle a combinatorial optimization problem on an Ising machine, the problem must be formulated as a quadratic unconstrained binary optimization (QUBO) model.
The QUBO model is given as
\begin{equation}
  \mathcal{H}_{\mathrm{QUBO}}(\{ \bm{x} \}) = \sum_{1\leq i \leq j \leq N}Q_{i, j}x_{i}x_{j},
  \label{eq:H_QUBO}
\end{equation}
where $x_{i} \in \{0, 1\}$ and $Q_{i, j}$ is the ($i$, $j$)-th element of the $N$-by-$N$ QUBO matrix $Q$.
Thus, a limitation of the Ising machines has been the need to manually construct a suitable QUBO of the target model.
However, in FMQA, this issue is handled by the mathematical equivalence between FM and QUBO.
The FM is defined by
\begin{equation}
    \mathcal{H}_{\rm{FM}}(\{ \bm{x} \}) = \omega_{0}+\sum_{i=1}^{N}\omega_{i}x_{i}+\sum_{1\le i < j \le N}\langle \bm{v}_i, \bm{v}_j \rangle x_{i}x_{j},
    \label{eq:H_FM}
\end{equation}
where $\omega_0 \in \mathbb{R}$, $\omega_{i} \in \mathbb{R}$, and $\bm{v}_{i} \in \mathbb{R}^K \ (i=1,\ldots,N)$ are the model parameters.
The parameter $K \in \mathbb{N}$ is the hyperparameter of the FM.
The symbol $\langle \cdot, \cdot \rangle$ denotes the inner product.
The FM model in Eq.~\eqref{eq:H_FM} can be converted into a QUBO model.
The diagonal elements of the $Q$ correspond to linear coefficients $\omega_{i}$ since $x_{i}^2 = x_{i}$ for any $i$, while the off-diagonal elements are given by $\langle \bm{v}_i, \bm{v}_j \rangle$.
Therefore, Ising machines can be applied even to problems where explicit QUBO modeling is difficult using FMQA.
FMQA has been successfully applied to various domains, including material design~\cite{kitai2020designing, nawa2023quantum, kim2024quantum, couzinie2025machine}, polymer design~\cite{tucs2023quantum, huang2024tutorial}, engineering design~\cite{matsumori2022application, inoue2022towards}, feature selection~\cite{tamura2024machine, kikuchi2026high}.
Previous studies have reported that it can obtain high-quality solutions with fewer evaluations compared to random search, genetic algorithms, particle swarm optimization, and Bayesian optimization, which is a representative BB optimization method~\cite{kitai2020designing, nawa2023quantum, huang2024tutorial, inoue2022towards}. 

In this study, we proposed a novel approach for applying FMQA to the RNA inverse folding problem.
To tackle the problem using FMQA, it is necessary to convert nucleotides, which are categorical variables, into binary variables.
Variables with more than two discrete states are commonly encoded as binary variables using binary-integer encoding.
In contrast, a previous study employed a binary variational autoencoder (bVAE) to learn binary latent representations of amino acids~\cite{tucs2023quantum}.
The advantage of bVAE is its ability to efficiently convert a vast variety of variables, such as amino acids, while reducing the total number of required binary variables.
However, since our study focuses on only four types of nucleotides, conventional binary-integer encoding is sufficient.
Several types of binary-integer encoding exist, and a previous study has reported that the choice of encoding can significantly influence the performance of FMQA~\cite{seki2022black}. 
A previous study has demonstrated that an appropriate mapping to binary variables can reshape the energy landscape, reduce the probability of being trapped in local minima, and thereby improve the solution quality of FMQA~\cite{koshikawa2025efficient}.
It is also known that the specific encoding method affects the solution accuracy when solving problems with Ising machines~\cite{chancellor2019domain, tamura2021performance, chen2021performance, kikuchi2024performance}. 
While a previous report has analyzed the effects of converting ordered integers into binary variables~\cite{seki2022black}, there are no studies that analyze the role of binary-integer encoding when handling categorical variables within FMQA.
Because nucleotides are categorical variables, their assignment to an integer is inherently arbitrary, and different binary-integer encoding methods may introduce distinct search biases. 
Therefore, this study investigates how binary-integer encodings and integer-to-nucleotide assignments influence solution quality in FMQA.
Furthermore, we aim to evaluate the applicability of FMQA to the RNA inverse folding problem.
The insights obtained here provide practical guidelines for applying FMQA to categorical optimization problems and lay the methodological foundation for its use in realistic RNA inverse folding scenarios.

\section*{Results}
\subsection*{Objective function of RNA inverse folding problem}
The RNA inverse folding problem aims to identify an RNA sequence that preferentially adopts a target structure.
In practical applications, the quality of an RNA sequence designed for a target secondary structure is evaluated by experimental validation.
However, to evaluate the proposed method, we use the ensemble defect as a computationally evaluable objective function in this study.

The ensemble defect quantifies the expected number of nucleotides whose pairing status differs from that of the target secondary structure over the Boltzmann ensemble of RNA secondary structures.
This metric evaluates not only the agreement between the MFE structure and the target structure, but also the thermodynamic stability of the target structure within the ensemble.
Since its introduction by Dirks et al.~\cite{dirks2004paradigms}, ensemble defect has been adopted in several subsequent studies~\cite{garcia2013rnaifold, zadeh2011nucleic, zadeh2011nupack}.
Dirks et al. demonstrated that ensemble defect outperforms structure distance based on MFE, achieving higher success rates in obtaining sequences that uniquely fold into the target structure~\cite{dirks2004paradigms}.
More recently, Ward et al. confirmed that ensemble defect consistently performs better than structure distance between the MFE structure and target structure across diverse benchmarks~\cite{ward2023fitness}.
Therefore, we used the ensemble defect.

An RNA sequence $\bm{n}$ of length $L$ is represented by a string of nucleotides $n_{i} \in \{$ A, U, G, C $\}$.
The secondary structure $s$ of $\bm{n}$ is defined as a set of paired positions $(i, j) \ (i < j)$ such that $(n_{i}, n_{j})$ forms an allowed base-pair type (A-U, G-C, or U-G).
Although natural RNAs may contain pseudoknots or higher-order interactions, these structural features are not supported by the folding model used here. 
Therefore, we consider only pseudoknot-free secondary structures.

Under the nearest-neighbor thermodynamic model~\cite{turner2010nndb}, each structure $s$ of $\bm{n}$ is assigned a Gibbs free energy $\Delta G(\bm{n}, s)$.
Assuming thermodynamic equilibrium at temperature $T$, the probability of observing $s$
follows the Boltzmann distribution
\begin{equation}
    p(s \ | \ \bm{n}) =
    \frac{\exp\!\left(-\Delta G(\bm{n}, s)/(RT)\right)}
    {Z(\bm{n})},
\end{equation}
where $R$ is the molar gas constant and
\begin{equation}
    Z(\bm{n}) =
    \sum_{s \in S(\bm{n})}
    \exp\!\left(-\Delta G(\bm{n}, s)/(RT)\right)
\end{equation}
is the partition function, with $S(\bm{n})$ denoting the ensemble of all
secondary structures that $\bm{n}$ can possibly fold into.

The ensemble defect is defined as the Boltzmann-weighted average structural distance
\begin{equation}
    \phi(\bm{n}, t) =
    \sum_{s \in S(\bm{n})}
    p(s \ | \ \bm{n})\, d(s, t),
\end{equation}
where $d(s, t)$ denotes the number of nucleotides whose pairing status differs between
structure $s$ and the target structure $t$.
The normalized ensemble defect (NED), which ranges between $0$ and $1$, is given by 
\begin{equation}
    \mathrm{NED}(\bm{n}, t) = \frac{\phi(\bm{n}, t)}{L}.
\end{equation}
The evaluation of ensemble defect was performed using the \verb+ViennaRNA+ package~\cite{lorenz2011viennarna}. 

\subsection*{Proposed method}
\begin{figure}[t]
    \centering
    \includegraphics[width=0.6\linewidth]{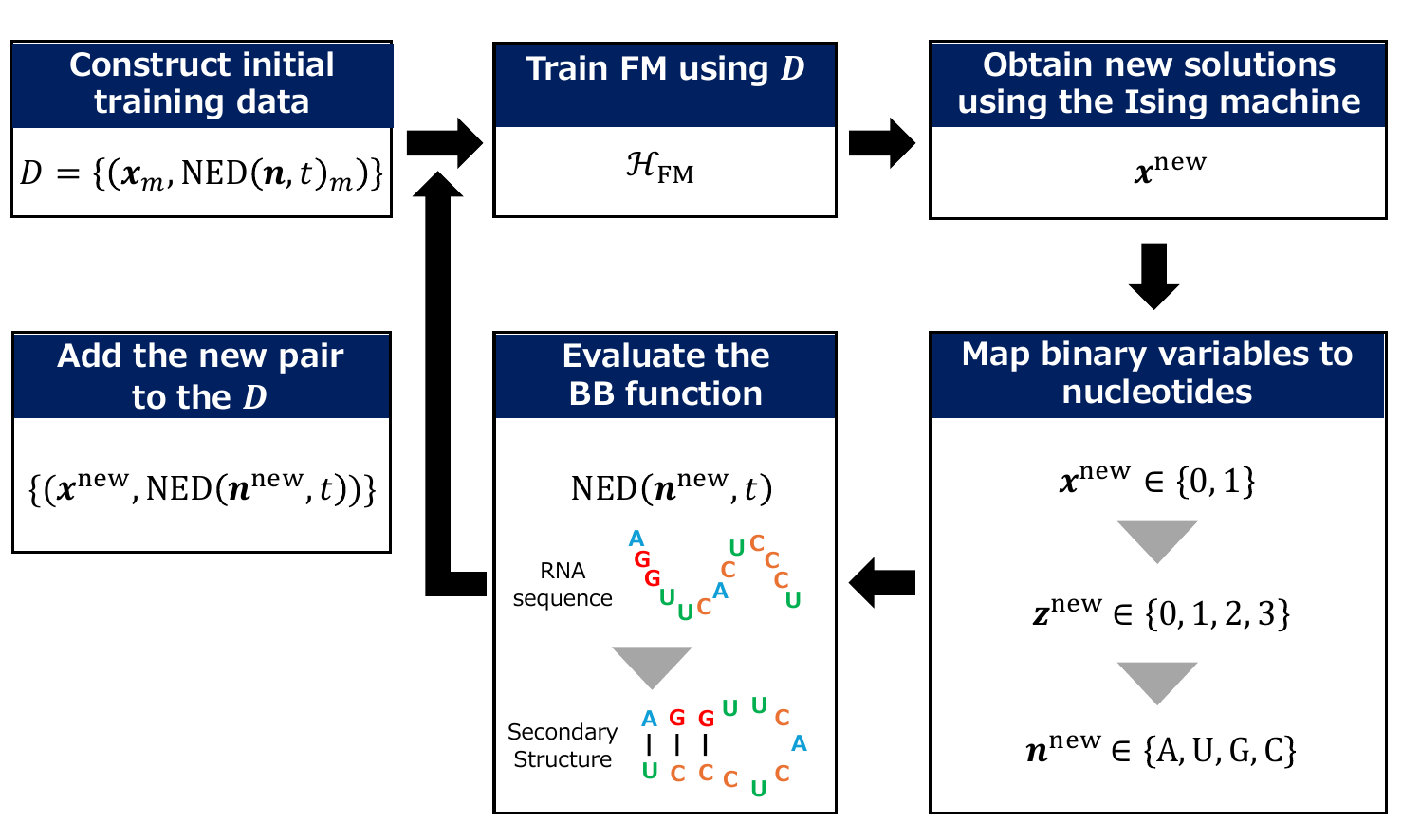}
    \caption{Schematic illustration of the proposed FMQA for the RNA inverse folding problem. As an example, the BB function is defined as the NED.}
    \label{fig:FMQA-RNA_procedure}
\end{figure}

We proposed a novel approach to solve the RNA inverse folding problem based on FMQA.
Figure~\ref{fig:FMQA-RNA_procedure} shows the procedure of the proposed method.

In practical applications, the BB function is typically assumed to correspond to experimentally measured properties, which may require substantial time and resources. 
However, to evaluate the applicability of FMQA in a controlled setting, we employed the computationally evaluable NED as the BB function in this study.

First, an initial training dataset $\{ (\bm{x}_{m}, \textrm{NED}(\bm{n}, t)_{m}) \} \ (m=1,\ldots,M)$ is constructed by randomly generating solutions of binary variables $\bm{x}_{m}$ and their corresponding $\textrm{NED}(\bm{n}, t)_{m}$.
We note that $M$ is the number of initial data points.
Here, the binary variables $\bm{x}$ are mapped integer variables $\bm{z}$ using binary-integer encoding, and then the integer variables are mapped to nucleotides $\bm{n}$.
Let $D$ denote the dataset constructed from these data points.

Next, an FM is trained using the dataset $D$. 
The FM model is defined in Eq.~\eqref{eq:H_FM}.
After training, new candidate solutions $\bm{x}^\textrm{new}$ are generated by optimizing an acquisition function using an Ising machine.
In this study, the predicted cost output by the trained FM is directly employed as the acquisition function.
The candidate solutions $\bm{x}^\textrm{new}$ are mapped to strings that represent the corresponding nucleotides $\bm{n}^\textrm{new}$.
Then, the $\textrm{NED}(\bm{n}^\textrm{new}, t)$ are evaluated using the candidate solutions.
The new pairs of solutions and their corresponding NEDs are appended to the dataset $D$.

This procedure, consisting of FM training, optimization using the Ising machine, and BB function evaluation, is repeated for a predefined number of iterations.
Finally, the solution with the lowest ensemble defect observed in the dataset $D$ is selected as the best solution obtained by FMQA.

\subsection*{Binary-integer encoding method}
In this study, we use four types of binary-integer encoding methods: one-hot, domain-wall, binary, and unary encoding.
Since RNA consists of four types of nucleotides, each nucleotide can be represented as a discrete integer variable with four possible states.
Table~\ref{tab:binary_integer_encoding} presents the binary variable sequences corresponding to each integer $I \in \{$ 0, 1, 2, 3 $\}$ for the different binary-integer encoding methods.
This subsection explains the binary representations of integers for each encoding method and their integration into FMQA.
\begin{table}[t]
    \centering
    \begin{tabular}{|c|c|c|c|c|}
    \hline
    \multirow{3}{*}{Integer} &
    \multicolumn{4}{c|}{Binary variable sequence} \\
    \cline{2-5}
    & \multicolumn{1}{c|}{\raisebox{-1.3em}{\shortstack[c]{One-hot \\ encoding}}}
    & \multicolumn{1}{c|}{\raisebox{-1.3em}{\shortstack[c]{Domain-wall \\ encoding}}}
    & \multicolumn{1}{c|}{\raisebox{-1.4em}{\shortstack[c]{Binary \\ encoding}}}
    & \multicolumn{1}{c|}{\raisebox{-1.4em}{\shortstack[c]{Unary \\ encoding}}} \\
    \hline
    0 & 1000 & 000 & 00 & 000 \\ \hline
    1 & 0100 & 100 & 10 & 100, 010, 001 \\ \hline
    2 & 0010 & 110 & 01 & 110, 101, 011 \\ \hline
    3 & 0001 & 111 & 11 & 111 \\ 
    \hline
    \end{tabular}
    \caption{Binary variable sequence representing integers $I \in \{0, 1, 2, 3\}$ under different binary-integer encoding methods. }
    \label{tab:binary_integer_encoding}
\end{table}

One-hot encoding is a traditional method.
An integer is represented by the position of a single binary variable that takes the value $1$.
An integer $I$ is given by $I = \sum_{i = 0}^{3} i x_{i}$.
To handle the one-hot encoding into FMQA, a penalty term is added to the FM model in Eq.~\eqref{eq:H_FM}.
The model is given as
\begin{equation}
    \mathcal{H}_{\mathrm{FM}}^{\mathrm{oh}} = \mathcal{H}_{\mathrm{FM}} + \mu \sum_{l = 1}^{L} \left( \sum_{i = 0}^3 x_{i}^{(l)} - 1 \right) ^ 2,
    \label{eq:OH_FM}
\end{equation}
where $\mu > 0$ is a penalty coefficient that controls the strength of the penalty term, and $x_{i}^{(l)}$ represents binary variables for an $l$-th integer variable.
Domain-wall encoding was originally proposed by Chancellor~\cite{chancellor2019domain}.
An integer is represented by the number of leading binary variables that take the value $1$, followed by $0$s.
An integer is given by $I = \sum_{i = 0}^{2} x_{i}$.
The boundary position between the binary variables set to $1$ and those set to $0$ is referred to as a domain wall.
To enforce the constraint that exactly one domain wall exists, a penalty term is added to the FM model.
The extended FM model is given as
\begin{equation}
    \mathcal{H}_{\mathrm{FM}}^{\mathrm{dw}} = \mathcal{H}_{\mathrm{FM}} + \mu \sum_{l = 1}^{L} \sum_{i = 0}^1 x_{i+1}^{(l)} (1 - x_{i}^{(l)}).
    \label{eq:DW_FM}
\end{equation}
Binary encoding uses the binary representation for encoding integers.
An integer $I$ is given by $I = \sum_{i = 0}^{1} 2^{i} x_{i}$.
In the RNA inverse folding problem, binary encoding does not require an additional penalty term, as each binary configuration uniquely corresponds to an integer value.
Unary encoding represents an integer by the number of binary variables set to value $1$, similar to domain-wall encoding.
Since this representation does not impose an explicit structural constraint, no penalty term is required.
Unary encoding admits multiple binary representations for a single integer value.
When applying FMQA, the total number of binary variables depends on the chosen encoding method.
One-hot encoding requires $N = 4L$, domain-wall and unary encodings require $N = 3L$, and binary encoding requires $N = 2L$.

A previous study investigated the effect of binary-integer encoding methods on FMQA.
One-hot, domain-wall, and binary encoding, were evaluated using the ground-state energy calculation of a hydrogen molecule~\cite{seki2022black}.
The problem was originally formulated with real-valued variables, and these variables were discretized using binary-integer encoding.
While all encoding methods achieved high accuracy, one-hot and domain-wall encodings outperformed binary encoding in terms of energy error distribution and stability.
In particular, one-hot encoding consistently yielded low energy errors even with small $K$.

\subsection*{Evaluation of effect of binary-integer encoding method and integer-to-nucleotide assignment}

\begin{figure}[t]
    \centering
    \includegraphics[width=0.95\linewidth]{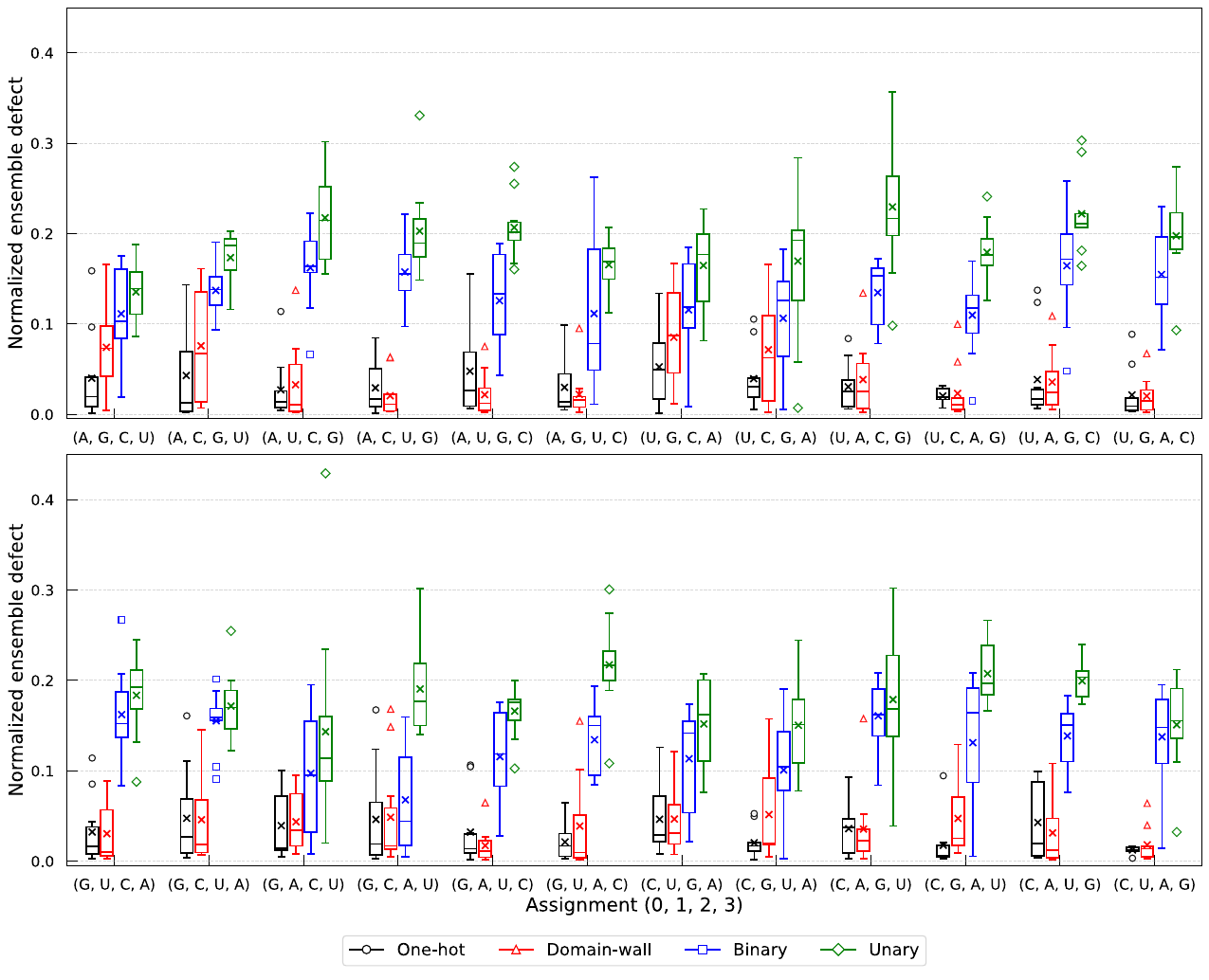}
    \caption{Normalized ensemble defect values obtained by FMQA under different combinations of binary-integer encoding methods and integer-to-nucleotide assignments. Crosses indicate the average NED over $10$ runs. The upper and lower whiskers denote the maximum and minimum NED values, respectively. Black circle, red triangle, blue square, and green diamond represent outliers. Panels in the top row correspond to assignments in which A or U was assigned to integer $0$, whereas panels in the bottom row correspond to assignments in which G or C was assigned to integer $0$.}
    \label{fig:ned_stickshift}
\end{figure}

To investigate the conditions under which FMQA can effectively solve the RNA inverse folding problem, we investigated the effects of binary-integer encoding and integer-to-nucleotide assignment.
For binary-integer encoding, we considered one-hot, domain-wall, binary, and unary encodings.
For integer-to-nucleotide assignment, we evaluated all possible $4! = 24$ assignments of the four nucleotides to the integers $(0,1,2,3)$.
For each condition, the optimization was performed independently $10$ times.

The objective function was the NED.
Since the NED represents the structural error with respect to the target structure, the problem is formulated as a minimization task.
The number of FMQA iterations was set to $1000$, and the number of initial training data points was set to $10$.
We used an SA-based Ising machine~\cite{FixAE}, which runs on a GPU.
Other parameters are described in the Methods section.
As the target structure, we used \textit{stickshift} from the well-known benchmark Eterna100~\cite{anderson2016principles}.
This structure consists of $26$ nucleotides (nt) and has a relatively simple secondary structure.
Thus, it provides an appropriate test case for evaluating the effects of encoding and assignment.

\begin{figure}[t]
    \centering
    \includegraphics[width=0.95\linewidth]{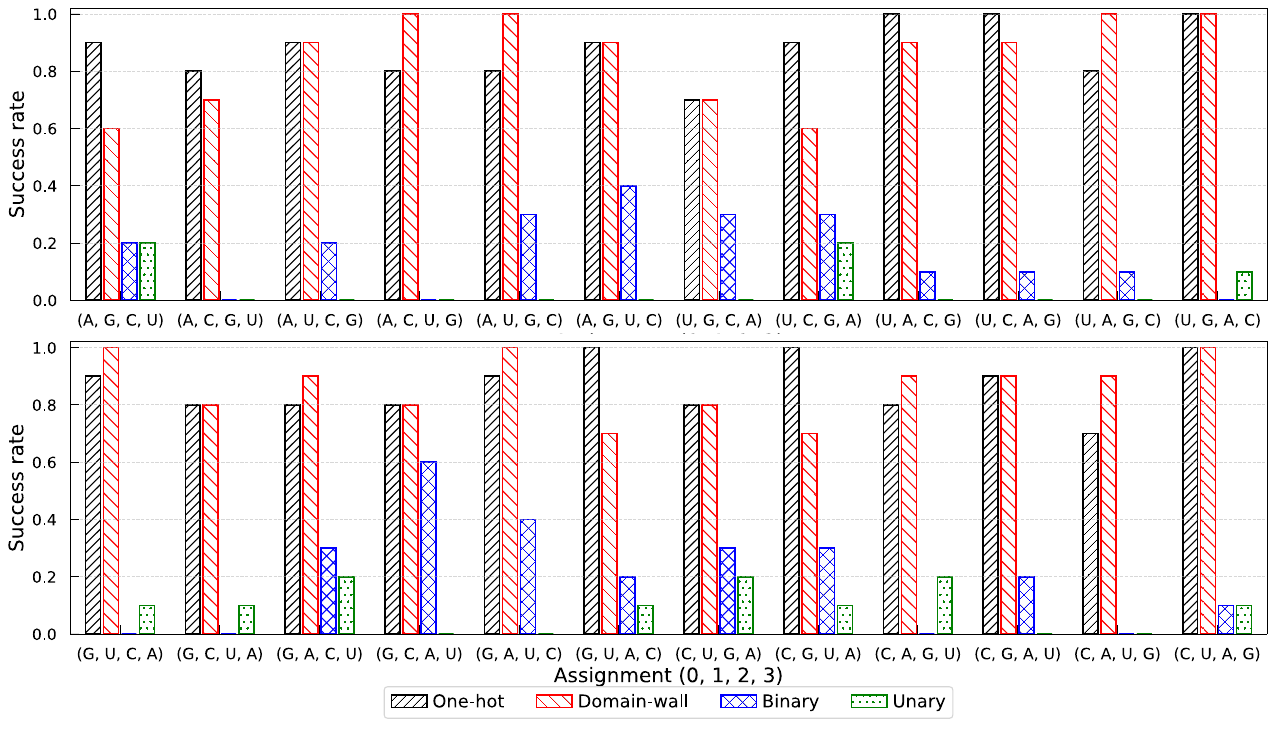}
    \caption{Success rate obtained by FMQA under different combinations of binary-integer encoding methods and integer-to-nucleotide assignments. Panels in the top row correspond to assignments in which A or U was assigned to integer $0$, whereas panels in the bottom row correspond to assignments in which G or C was assigned to integer $0$.}
    \label{fig:success_rate_stickshift}
\end{figure}

Figure~\ref{fig:ned_stickshift} shows the solutions obtained by FMQA under different combinations of binary-integer encoding methods and integer–to-nucleotide assignments. 
The ensemble defect corresponds to the expected number of nucleotides whose pairing status differs from that of the target secondary structure under the Boltzmann ensemble of possible structures.
Therefore, a lower NED indicates that the sequence more reliably adopts the target structure.
Focusing on the binary-integer encoding methods, one-hot and domain-wall encodings achieved lower NED values compared with binary and unary encodings. 
Binary encoding yielded lower NED values than unary encoding.
With respect to integer–to-nucleotide assignment, the ensemble defect values of domain-wall encoding were relatively high for the assignments (A, G, C, U), (A, C, G, U), (U, G, C, A), and (U, C, G, A). 
In contrast, one-hot encoding consistently achieved low NED values regardless of the assignment.

The objective of the RNA inverse folding problem is to identify RNA sequences whose MFE secondary structure coincides with the target structure. 
Therefore, we computed the MFE structure for each RNA sequence obtained by FMQA and examined whether it matched the target structure. 
Solutions whose MFE structures were identical to the target structure were defined as success solutions, and the proportion of success solutions among the $10$ runs was defined as the success rate. 
The success rates under each condition are shown in Fig.~\ref{fig:success_rate_stickshift}.

Binary and unary encodings exhibited relatively high NED values, and consequently, their success rates were low, with some conditions yielding a success rate of $0.0$.
In contrast, one-hot and domain-wall encodings generally achieved high success rates. 
However, for the assignments in which domain-wall encoding showed high NED values, the success rate was correspondingly low.

\begin{figure}[t]
    \centering
    \includegraphics[width=0.95\linewidth]{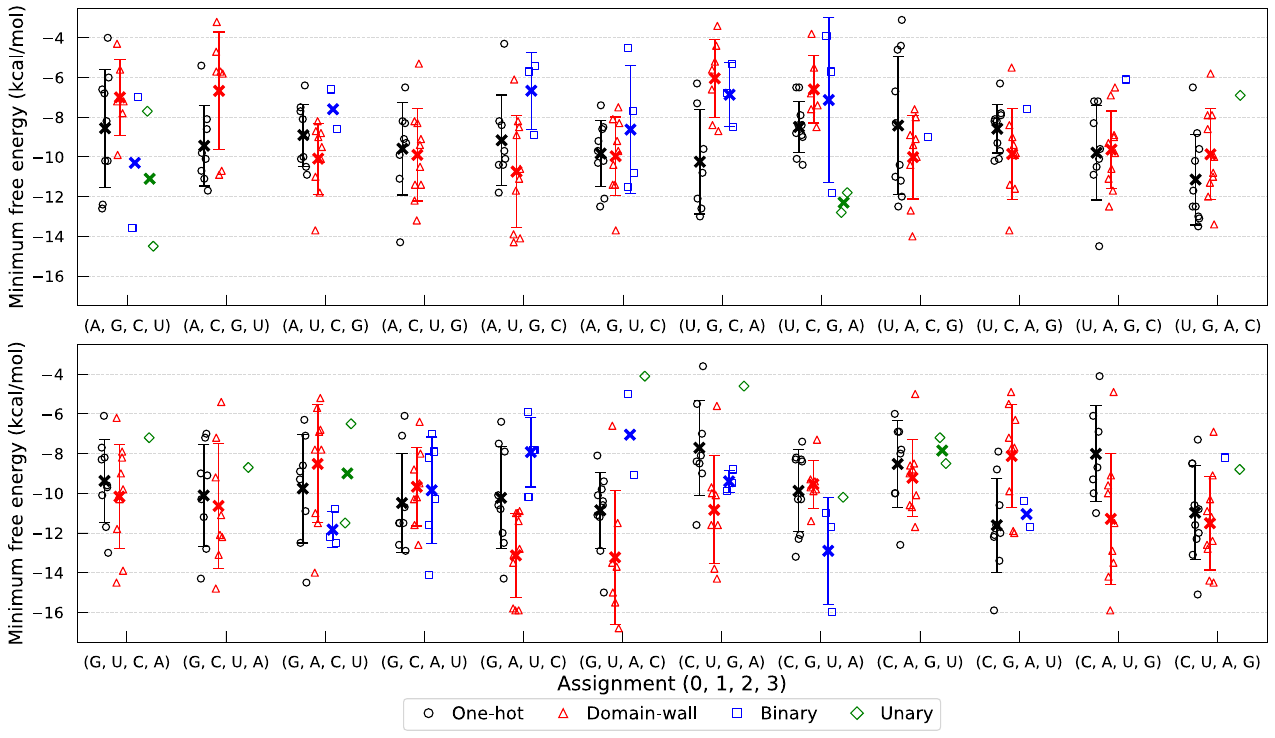}
    \caption{Minimum free energy values obtained by FMQA under different combinations of binary-integer encoding methods and integer-to-nucleotide assignments. Only success solutions are plotted. When two or more success solutions were obtained, their average MFE value is indicated by a cross marker. When three or more success solutions were obtained, the standard deviation is shown as error bars. Panels in the top row correspond to assignments in which A or U was assigned to integer $0$, whereas panels in the bottom row correspond to assignments in which G or C was assigned to integer $0$.}
    \label{fig:mfe_stickshift}
\end{figure}

Finally, we analyzed the MFE values of the success solutions. 
The MFE values of the success solutions are presented in Fig.~\ref{fig:mfe_stickshift}. 
One-hot encoding yielded solutions with comparable MFE values (approximately $-10$ kcal/mol) across all assignments.
In contrast, domain-wall encoding exhibited assignment-dependent behavior. 
For the assignments (A, G, C, U), (A, C, G, U), (U, G, C, A), and (U, C, G, A), where the NED values were high, the corresponding MFE values were also relatively high.
Conversely, for the assignments (G, A, U, C), (G, U, A, C), (C, A, U, G), and (C, U, A, G), the MFE values were lower compared to those obtained with other assignments.

\subsection*{Comparison of the number of BB function evaluations}
In the previous subsection, we analyzed the effects of binary-integer encoding and nucleotides assignments from the viewpoint of final solutions obtained by FMQA.
In this subsection, we compare FMQA with other black-box optimization methods in terms of the number of objective function evaluations.
In practical applications, each BB function evaluation may correspond to a wet-lab experiment, which typically requires a substantial amount of time. 
In such settings, the computational time of surrogate training and optimization is negligible compared to the time required for wet-lab experimentation. 
Therefore, the number of BB function evaluations is the primary performance metric.

Although several methods have been proposed for the RNA inverse folding problem~\cite{hofacker1994fast, andronescu2004new, busch2006info, zadeh2011nucleic, zadeh2011nupack, taneda2010modena, lyngso2012frnakenstein, rubio2018multiobjective, garcia2013rnaifold, kleinkauf2015antarna, yang2017rna}, their objective functions often differ from that used in our study, and in many cases the number of objective function evaluations cannot be directly measured or fairly compared. 
Therefore, in this study, we compared the number of evaluations using Bayesian optimization and a genetic algorithm (GA), both using NED as the objective function.

For Bayesian optimization, we employed the tree-structured Parzen estimator (TPE), which can directly handle categorical variables. 
For FMQA, we used the integer-to-nucleotide assignment (G, A, U, C), which achieved a success rate of $1.0$, low MFE values, and the lowest NED among the domain-wall encoding results in the previous subsection. 
The results obtained with one-hot and domain-wall encodings under this assignment were used for comparison.
In addition, random search (RS) was adopted as a baseline method. 
All methods were initialized with the same initial dataset as FMQA to ensure a fair comparison.

\begin{figure}[t]
    \centering
    \includegraphics[width=0.45\linewidth]{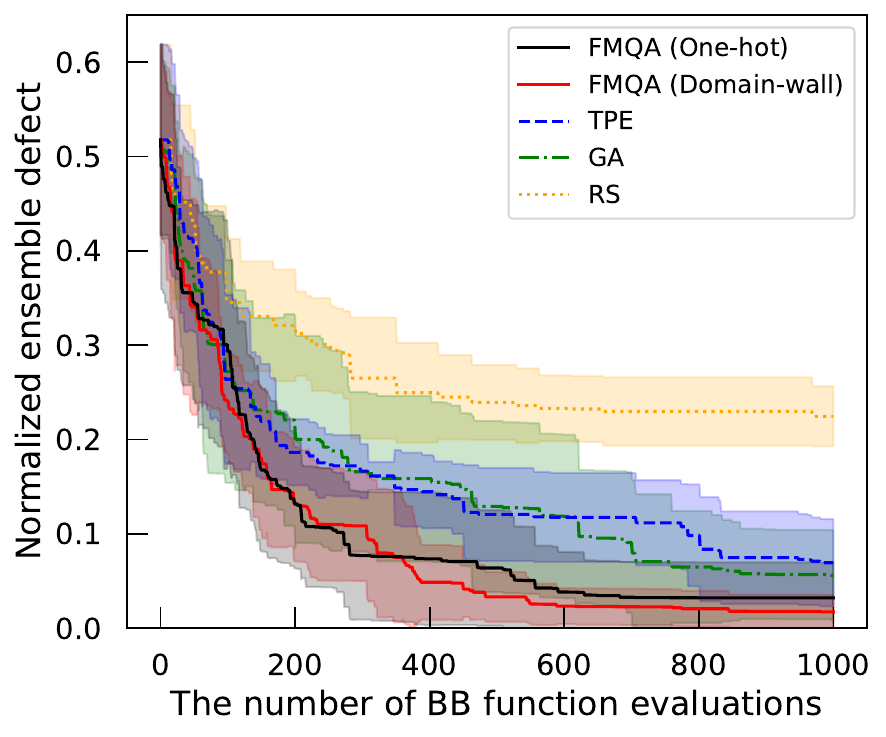}
    \caption{Relationship between the number of BB function evaluations and the best normalized ensemble defect value up to each evaluation.}
    \label{fig:comparison_number}
\end{figure}

Figure~\ref{fig:comparison_number} shows the relationship between the number of NED evaluations and the best NED value obtained up to each evaluation for each method. 
Each method was independently performed $10$ times, and the mean and standard deviation across runs are presented.
As shown in Fig.~\ref{fig:comparison_number}, FMQA achieved lower NED values with fewer evaluations compared to the other methods.
The result demonstrates its efficiency in reducing the number of expensive objective function evaluations in the RNA inverse folding problem.

\subsection*{Performance on multiple target secondary structures}
In this subsection, we evaluated the performance of FMQA on target secondary structures other than \textit{stickshift}.

\begin{figure}[t]
    \centering
    \includegraphics[width=0.90\linewidth]{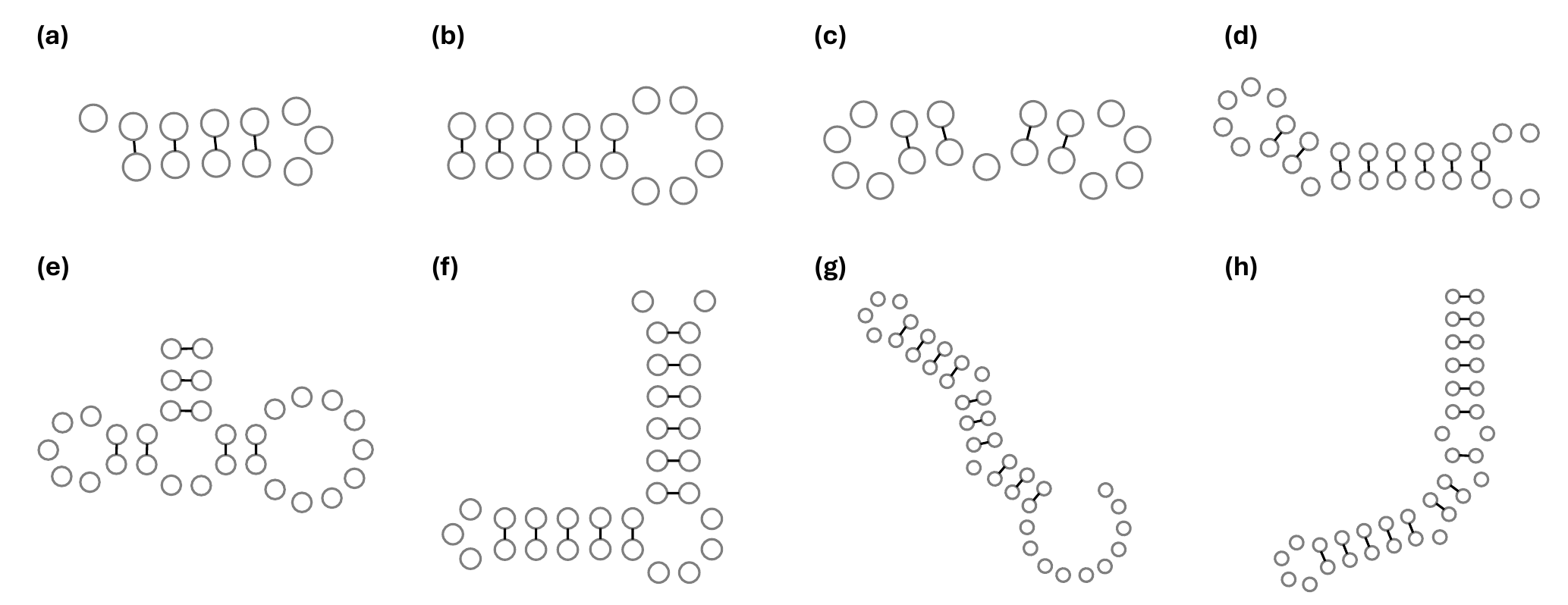}
    \caption{Target secondary structures selected from the Eterna100 benchmark and evaluated in this study. White circles represent nucleotides, and black lines indicate base pairs formed by hydrogen bonding. The target structures are: (a) \textit{G-C Placement} ($12$ nt), (b) \textit{Simple Hairpin} ($16$ nt), (c) \textit{Shortie 4} ($17$ nt), (d) \textit{stickshift} ($26$ nt), (e) \textit{Small and Easy 6} ($30$ nt), (f) \textit{Corner bulge training} ($31$ nt), (g) \textit{Prion Pseudoknot – Difficulty Level 0} ($36$ nt), and (h) \textit{infoRNA test 16} ($36$ nt).}
    \label{fig:selected_structure}
\end{figure}

From the Eterna100 benchmark, we selected eight structures with relatively small nucleotide lengths (12–36 nt). 
The selected target structures are shown in Fig.~\ref{fig:selected_structure}.

FMQA was performed using the integer-to-nucleotide assignment (G, A, U, C), which demonstrated better performance in the previous subsection. 
We compared one-hot and domain-wall encoding under the same experimental conditions. 
For each target structure, the optimization was independently executed $10$ times.

\begin{figure}[t]
    \centering
    \includegraphics[width=0.65\linewidth]{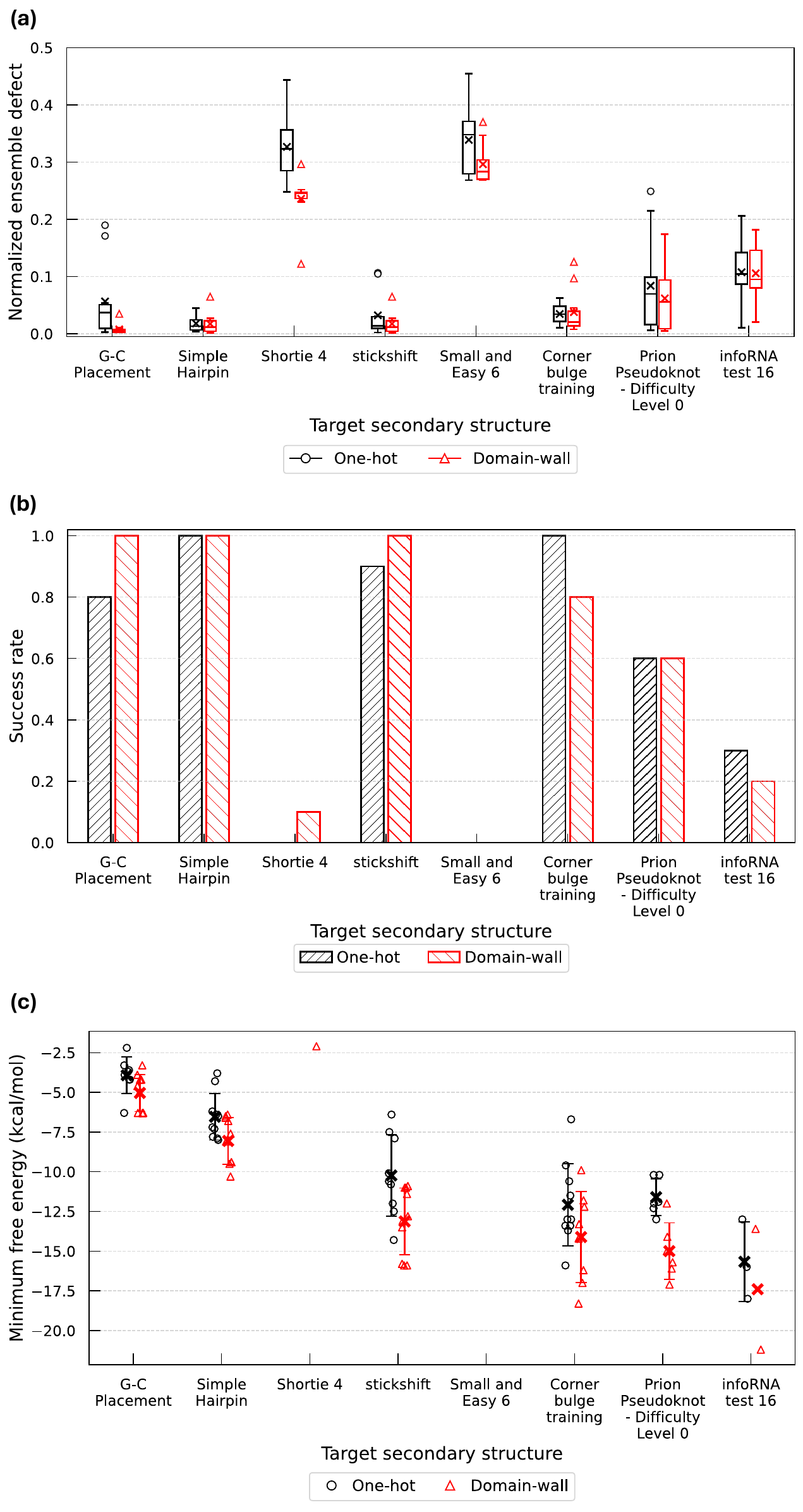}
    \caption{Performance metrics on multiple Eterna100 benchmark target structures.
    (a) Normalized ensemble defect values obtained by FMQA for each target structure. Crosses indicate the average NED over $10$ independent runs. The upper and lower whiskers denote the maximum and minimum NED values, respectively. Black circle and red triangle represent outliers. (b) Success rate. (c) Minimum free energy values. Only success solutions are plotted. When two or more success solutions were obtained, their average MFE value is indicated by a cross marker. When three or more success solutions were obtained, the standard deviation is shown as error bars.}
    \label{fig:comparison_target}
\end{figure}

Figure~\ref{fig:comparison_target}(a) presents the NED obtained for each structure.
Overall, domain-wall encoding achieved lower NED values than one-hot encoding for most target structures. 
In particular, for \textit{G-C Placement}, \textit{Simple Hairpin}, \textit{stickshift}, and \textit{Corner bulge training}, both encoding methods consistently yielded low NED values across runs. 
In contrast, \textit{Shortie 4} and \textit{Small and Easy 6} showed relatively higher NED values.
The corresponding success rates are shown in Fig.~\ref{fig:comparison_target}(b).
Except for \textit{Shortie 4} and \textit{Small and Easy 6}, at least two success solutions were obtained for each structure using both encoding methods.
Notably, for \textit{G-C Placement, Simple Hairpin, stickshift}, and \textit{Corner bulge training}, the success rate was high for both encodings.
In contrast, \textit{Prion Pseudoknot – Difficulty Level 0} and \textit{infoRNA test 16} exhibited lower success rates compared with their target structures.
Figure~\ref{fig:comparison_target}(c) shows the MFE values of the success solutions. 
Consistent with the NED, domain-wall encoding achieved lower MFE values than one-hot encoding for most target structures. 

\section*{Discussion}
In this study, we demonstrated that, in the RNA inverse folding problem solved by FMQA, one-hot and domain-wall encodings achieved better solutions than binary and unary encodings. 
This is consistent with a previous study that investigated the effect of binary-integer encoding methods on FMQA when continuous variables were discretized into binary variables~\cite{seki2022black}.
In that study, the energy error, defined as the difference between the optimal objective value and the value obtained by FMQA, was smaller for one-hot and domain-wall encodings than for binary encoding.
In contrast, a previous study on the relationship between Ising machines and binary-integer encodings reported that binary and unary encodings sometimes outperformed one-hot encoding in terms of feasibility rate and objective value quality~\cite{tamura2021performance}.
These findings suggest that, in FMQA, the choice of binary-integer encoding primarily influences solution quality through the construction and expressivity of the FM, rather than through the optimization performance of the Ising machine.
Binary encoding represents four nucleotides using only two binary variables.
Although this representation is compact, it may limit the expressive capacity of the FM to model complex nonlinear interactions among categorical states.
For unary encoding, multiple binary configurations correspond to the same integer value.
This redundancy increases representational degeneracy. 
The FM must implicitly learn that these distinct binary configurations correspond to the same nucleotide, which may complicate surrogate modeling and reduce optimization efficiency.
Therefore, the observed degradation in solution quality for unary encoding may be attributed to the inability of the FM to properly capture this many-to-one mapping structure.

\begin{figure}[t]
    \centering
    \includegraphics[width=0.9\linewidth]{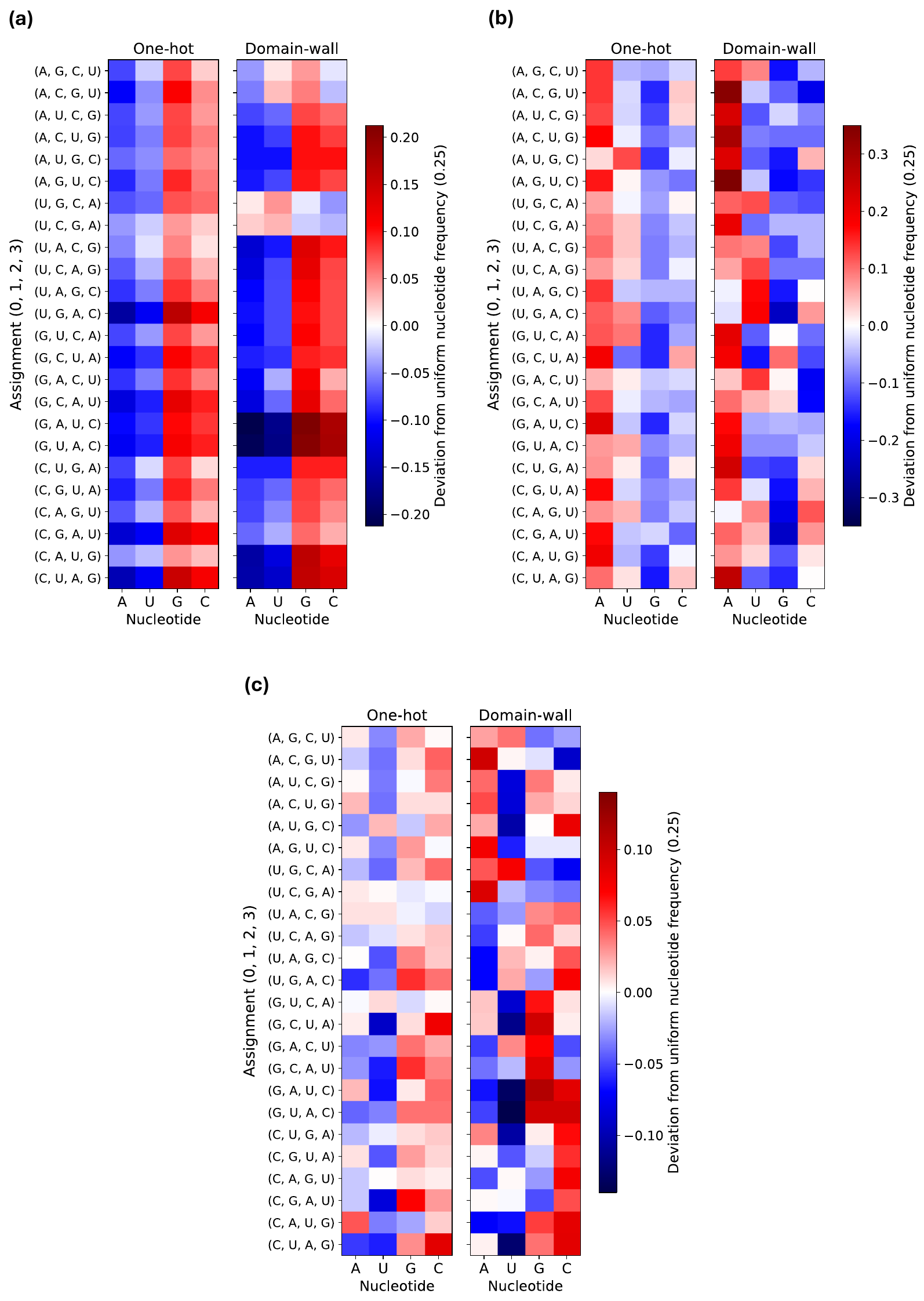}
    \caption{Nucleotide frequencies obtained from the success solutions across different integer-to-nucleotide assignments. (a) Nucleotide frequency in stem regions, (b) nucleotide frequency in non-stem regions, and (c) nucleotide frequency over the entire RNA sequence. The baseline value of 0.25 corresponds to the expected frequency under a random uniform nucleotide frequency. Red indicates frequencies higher than 0.25, whereas blue indicates frequencies lower than 0.25.}
    \label{fig:nucleotide_frequency}
\end{figure}

We next discuss the effect of integer–to-nucleotide assignment.
In domain-wall encoding, certain assignments resulted in different NED and MFE values. 
To analyze these results, we investigated nucleotide frequency in the solutions obtained under all assignments.
Based on the secondary structure, nucleotides were categorized into stem regions (base-paired positions) and non-stem regions (unpaired positions).
Nucleotide frequencies were calculated by pooling all nucleotides from the successful solutions obtained in $10$ independent runs and normalizing by the total number of nucleotides.
The results are shown in Fig.~\ref{fig:nucleotide_frequency}. 
Figure~\ref{fig:nucleotide_frequency}(a) shows nucleotide frequency in stem regions, Fig.~\ref{fig:nucleotide_frequency}(b) in non-stem regions, and Fig.~\ref{fig:nucleotide_frequency}(c) across the entire sequence. 
The baseline value of $0.25$ corresponds to uniform nucleotide frequency.
The heatmap shows the deviation of nucleotide frequency from this uniform expectation.

In stem regions(Fig.~\ref{fig:nucleotide_frequency}(a)), one-hot encoding consistently exhibited high G and C frequency across all assignments. 
This observation agrees with previous reports on RNA inverse folding~\cite{andronescu2004new, dirks2004paradigms, busch2006info, rubio2018multiobjective}. 
G-C base pairs form three hydrogen bonds and are typically more stable than A-U or U-G pairs.
The stability of the structure is determined by nearest-neighbor stacking interactions between adjacent base pairs.
In the Turner thermodynamic model implemented in the \verb+ViennaRNA+ package, G-C stacks exhibit the most favorable free energies, explaining why enrichment of G-C pairs in stem regions lowers MFE and reduces ensemble defect~\cite{mathews2004incorporating, turner2010nndb}.
In contrast, domain-wall encoding exhibited assignment-dependent results.
In assignments where A-U and U-G pairs were more prevalent in stems, NED and MFE values were higher: (A, G, C, U), (A, C, G, U), (U, G, C, A), and (U, C, G, A).
These assignments correspond to those that showed inferior optimization performance. 
Since A-U and U-G pairs are thermodynamically less stable than G-C pairs, enrichment of these pairs in stems increases free energy and reduces structural stability.
Meanwhile, assignments in which G-C frequency in stems exceeded that of one-hot encoding corresponded to lower MFE values, (G, A, U, C), (G, U, A, C), (C, A, U, G), and (C, U, A, G). 
This further supports the interpretation that stem thermodynamic stability directly influences FMQA performance in RNA inverse folding.

In non-stem regions(Fig.~\ref{fig:nucleotide_frequency}(b)), one-hot encoding showed a high frequency of adenine, consistent with the results of previous methods~\cite{rubio2018multiobjective}. 
Adenine forms canonical base pairs only with uracil. 
Compared with guanine, which can pair with both cytosine and uracil, adenine reduces the likelihood of unintended base pairing. 
Although cytosine also pairs with only guanine, erroneous C-G pairing is energetically more stabilizing than A-U pairing. 
Therefore, placing adenine in loop regions reduces the risk of forming undesired stable alternative structures, thereby lowering the ensemble defect.
In domain-wall encoding, adenine frequency remained relatively high in non-stem regions.
However, the frequency of other nucleotides varied substantially depending on assignment. 
Notably, nucleotides assigned to integer $0$ tended to exhibit higher frequency.

When examining overall nucleotide frequency (Fig.~\ref{fig:nucleotide_frequency}(c)), domain-wall encoding showed increased frequency for nucleotides assigned to integers $0$ and $3$.
This phenomenon can be interpreted from the viewpoint of Hamming distance between binary representations. 
In one-hot encoding, the Hamming distance between any pair of integers is uniformly two. 
In contrast, domain-wall encoding exhibits asymmetric distances.
Adjacent integers differ by Hamming distance one, whereas non-adjacent integers differ by two or three binary variables.
Adjacent integers are directly connected within the feasible subspace, while non-adjacent integers are separated by larger Hamming distances~\cite{kikuchi2024performance}.
In domain-wall encoding, integers $0$ and $3$ each have only one neighboring integer, whereas middle integers ($1$ and $2$) have two such neighbors. 
This asymmetry may bias the search dynamics, making transitions away from $0$ and $3$ less frequent.
As a result, nucleotides assigned to boundary integers $0$ and $3$ may appear more frequently in the optimized sequences.

In the present study, we did not explicitly constrain GC content or impose thermodynamic priors beyond ensemble defect minimization. 
Therefore, achieving low free energy is central to success in the RNA inverse folding problem considered here.
The results suggest that, when using domain-wall encoding, careful design of integer–to-nucleotide assignment can exploit encoding-specific search biases. 
For example, assigning G and C to integers whose binary representations are less prone to transition may enhance stem stability and reduce free energy.
Thus, domain-wall encoding provides not only a representation method but also an opportunity for encoding-aware problem design.

Finally, we discuss the influence of target secondary structure on FMQA performance.
In the RNA inverse folding problem, it has been reported that structures containing very short stems (e.g., two base pairs) flanked by unpaired nucleotides are particularly difficult to design~\cite{anderson2016principles}.
Such short stems are thermodynamically unstable and can be easily disrupted by alternative competing structures, making it difficult to identify sequences that uniquely fold into the target structure.
Among the target structures evaluated in this study, \textit{Shortie 4} and \textit{Small and Easy 6}, for which almost no success solutions were obtained, exhibit this characteristic feature. 
Our results indicate that FMQA also struggles with these inherently unstable structural motifs.
In contrast, bulge-loop structures, where one strand of a continuous helix contains unpaired nucleotides, are also known to increase design difficulty because they introduce asymmetry and local flexibility. 
Nevertheless, FMQA successfully obtained solutions for \textit{Corner bulge training}, which contains such bulge motifs.
This result suggests that FMQA can effectively handle certain classes of structural irregularities, provided that the overall stem length remains sufficiently long to ensure thermodynamic stability.
Furthermore, even when a structure contains multiple stable stems, increasing the total nucleotide length expands the combinatorial search space exponentially. 
As sequence length increases, the number of possible nucleotide combinations grows as $4^L$, thereby increasing the complexity of the optimization landscape. 
Consequently, the $36$-nt structures (\textit{Prion Pseudoknot – Difficulty Level 0} and \textit{infoRNA test 16}) exhibited lower success rates compared with shorter targets. 
This reduction in performance can be attributed not only to structural complexity but also to the enlarged search space associated with longer sequences.
These observations indicate that FMQA performance is influenced by both the thermodynamic stability of local motifs and global combinatorial complexity.
Addressing these difficulties may require the introduction of explicit base-pairing constraints or the adoption of motif-level decomposition strategies explored in previous RNA inverse folding studies~\cite{andronescu2004new, dirks2004paradigms}.
Future research will focus on extending the applicability of FMQA to a broader class of RNA inverse folding problems and systematically comparing its performance with established approaches~\cite{hofacker1994fast, andronescu2004new, busch2006info, zadeh2011nucleic, zadeh2011nupack, taneda2010modena, lyngso2012frnakenstein, rubio2018multiobjective, garcia2013rnaifold, kleinkauf2015antarna, yang2017rna, shi2018sentrna, koodli2019eternabrain, eastman2018solving, runge2018learning, sumi2024deep, akiyama2022informative}.

\section*{Methods}
\subsection*{RNA secondary structure analysis}
RNA secondary structure prediction and ensemble evaluation were performed using the \verb+ViennaRNA+ package (v2.7.2)~\cite{lorenz2011viennarna}. 
All calculations were conducted at a constant temperature of 37$^\circ$C to reflect physiological conditions.
This temperature corresponds to the default setting of the package and has been widely adopted in previous RNA folding studies~\cite{hofacker1994fast, zadeh2011nupack, rubio2018multiobjective, ward2023fitness}.
For the energy model, the dangle treatment was set to 2 (\texttt{--dangles=2}). 
This setting accounts for the energetic contributions of dangling ends and terminal mismatches, approximating coaxial stacking effects and providing a thermodynamically consistent estimation of structural stability.

\subsection*{FMQA settings}
\begin{figure}[t]
    \centering
    \includegraphics[width=0.95\linewidth]{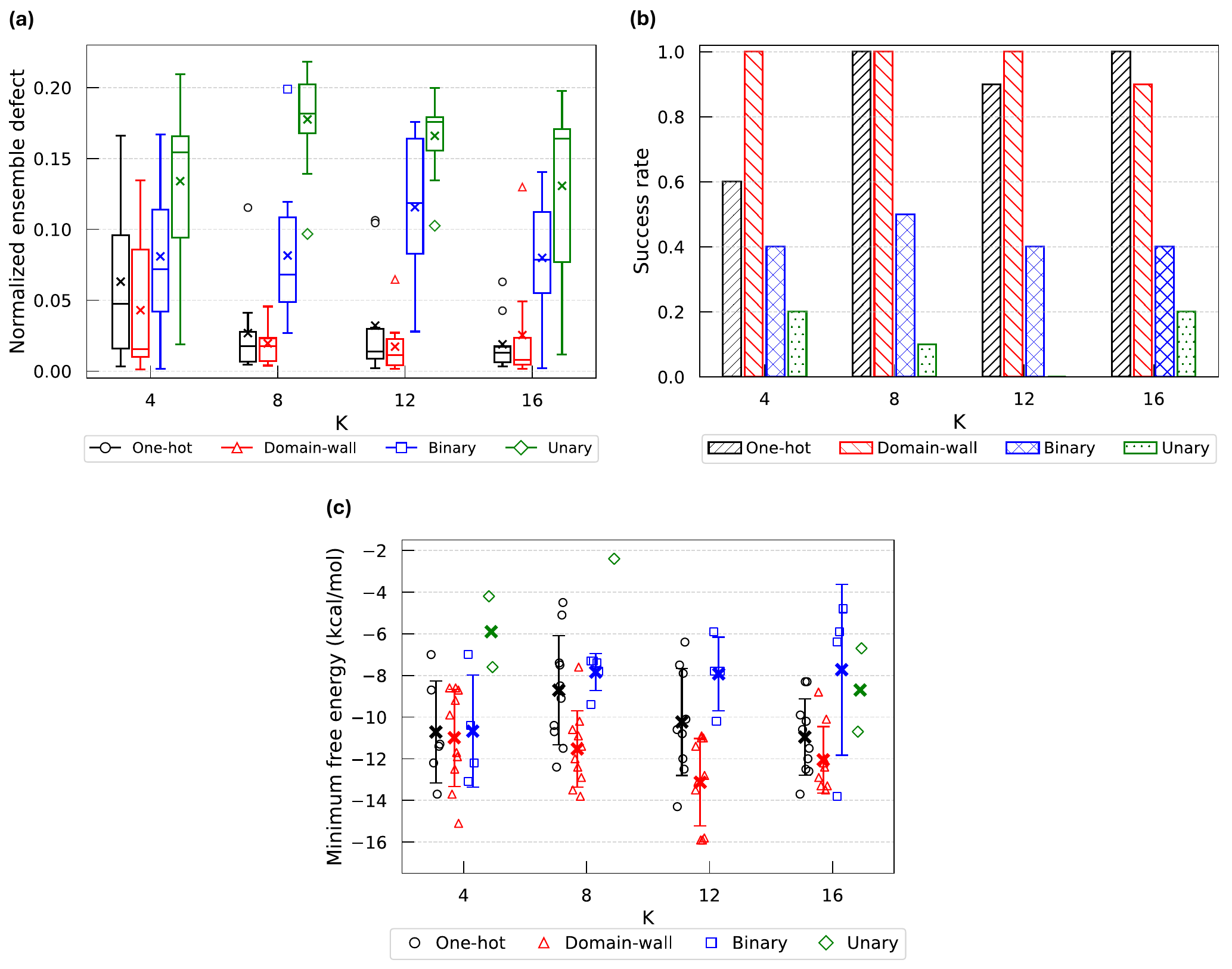}
    \caption{Performance metrics for different values of the hyperparameter $K$ under each binary-integer encoding method.
    (a) Normalized ensemble defect values obtained by FMQA for each encoding method and value of $K$. Crosses indicate the average NED over $10$ independent runs. The upper and lower whiskers denote the maximum and minimum NED values, respectively. Black circle, red triangle, blue square, and green diamond represent outliers. (b) Success rate. (c) Minimum free energy values of the success solutions. Only success solutions are plotted. When two or more success solutions were obtained, their average MFE value is indicated by a cross marker. When three or more success solutions were obtained, the standard deviation is shown as error bars.}
    \label{fig:comparison_k}
\end{figure}

The FM was trained using the AdamW optimizer with a learning rate of $0.01$. 
The number of training epochs was set to $1000$, and the mean squared error (MSE) was used as the loss function.
The hyperparameter $K$ was determined based on a comparative performance analysis. 
We evaluated $K = 4, 8, 12,$ and $16$ for each binary-integer encoding method. 
The integer-to-nucleotide assignment was fixed to (G, A, U, C). 
For each condition, the optimization was independently performed $10$ times. 
The results are shown in Fig.~\ref{fig:comparison_k}.
Figure~\ref{fig:comparison_k}(a) presents the NED values. 
One-hot and domain-wall encodings exhibited low NED values for all $K$ except $K = 4$. 
Binary encoding showed slight performance degradation at $K = 12$, and unary encoding at $K = 8$. Across all $K$, the order of performance was consistent: one-hot and domain-wall encodings achieved the lowest NED values, followed by binary encoding, and finally unary encoding.
The success rates are shown in Fig.~\ref{fig:comparison_k}(b). 
One-hot encoding exhibited a lower success rate at $K = 4$, whereas domain-wall encoding maintained a high success rate across all $K$.
Figure~\ref{fig:comparison_k}(c) shows the MFE values of the obtained success solutions. 
For one-hot encoding, the MFE values at $K = 8$ were slightly higher than those at other $K$ values. In domain-wall encoding, the lowest MFE values were observed at $K = 12$.
Based on these results, and considering the performance of one-hot and domain-wall encodings, we selected $K = 12$ for the FMQA experiments reported in the Results section.

The trained FM model was solved using an SA-based Ising machine~\cite{FixAE} implemented on a GPU. 
The limit of computation time per optimization run was set to $2000$~ms.
Before solving the FM model on the Ising machine, all coefficients were normalized by dividing them by the maximum absolute coefficient value. 
For one-hot and domain-wall encodings, a penalty coefficient was set to $\mu = 2$.
During the FMQA iterations with this penalty coefficient, no infeasible solutions were observed.
At each FMQA iteration, the Ising machine generated one new candidate solution $\bm{x}^{\mathrm{new}}$, which was evaluated to evaluate the NED and appended to the dataset.

The initial dataset consisted of $10$ randomly generated sequences. 
Each sequence was constructed by randomly selecting nucleotides from the set $\{A, U, G, C\}$, with the sequence length equal to that of the target secondary structure.
FMQA was performed for $1000$ iterations. 
The best solution observed during the optimization process was selected as the final output of FMQA.

The above settings were kept identical across all target secondary structures.

\subsection*{Baseline optimization methods}

Random search (RS) and tree-structured Parzen estimator (TPE) were implemented using the \verb+Optuna+ package. 
For TPE, the multivariate parameter was set to \texttt{True} to enable multivariate sampling of categorical variables.

The genetic algorithm (GA) was implemented using the \verb+pymoo+ package~\cite{blank2020pymoo}, which has been used in previous comparative studies with FMQA~\cite{huang2024tutorial}. 
The GA was formulated as a single-objective optimization problem.
Integer variables representing nucleotides were optimized directly in the range $\{0,1,2,3\}$. 
The population size was set to $10$, identical to the initial dataset size used in FMQA to ensure fair comparison.
Uniform crossover was employed with crossover probability $0.9$. 
Mutation was implemented as a random-reset mutation operator, where each gene was mutated with probability $p = 5.0 / L$, with $L$ denoting the sequence length (for stickshift, $L = 26$). 
When mutation occurred, the nucleotide value was randomly reassigned to a different integer.
The termination criterion was based on the total number of objective function evaluations. 
The evaluation budget was set to $1000$ objective function evaluations, excluding the evaluations of the initial dataset, to match the number of additional NED evaluations performed in FMQA.

All methods were initialized with the same initial dataset as FMQA.

\bibliography{ref}

@article{kitai2020designing,
  title={Designing metamaterials with quantum annealing and factorization machines},
  author={Kitai, Koki and Guo, Jiang and Ju, Shenghong and Tanaka, Shu and Tsuda, Koji and Shiomi, Junichiro and Tamura, Ryo},
  journal={Phys. Rev. Res.},
  volume={2},
  issue={1},
  note={{Art.} no. 013319},
  year={2020},
  publisher={APS}
}

@inproceedings{rendle2010factorization,
  title={Factorization machines},
  author={Rendle, Steffen},
  booktitle={2010 IEEE International conference on data mining},
  pages={995--1000},
  year={2010}
}

@article{seki2022black,
  title={Black-box optimization for integer-variable problems using {I}sing machines and factorization machines},
  author={Seki, Yuya and Tamura, Ryo and Tanaka, Shu},
  journal={arXiv preprint arXiv:2209.01016},
  year={2022}
}

@article{kikuchi2026high,
  title={High-Order Epistasis Detection Using Factorization Machine with Quadratic Optimization Annealing and {MDR}-Based Evaluation},
  author={Kikuchi, Shuta and Tanaka, Shu},
  journal={arXiv preprint arXiv:2602.01860},
  year={2026}
}

@article{matsumori2022application,
  title={Application of {QUBO} solver using black-box optimization to structural design for resonance avoidance},
  author={Matsumori, Tadayoshi and Taki, Masato and Kadowaki, Tadashi},
  journal={Sci. Rep.},
  volume={12},
  note={{Art.} no. 12143},
  year={2022},
  publisher={Nature Publishing Group UK London}
}

@article{tamura2025black,
  title={Black-box optimization using factorization and Ising machines},
  author={Tamura, Ryo and Seki, Yuya and Minamoto, Yuki and Kitai, Koki and Matsuda, Yoshiki and Tanaka, Shu and Tsuda, Koji},
  journal={arXiv preprint arXiv:2507.18003},
  year={2025}
}

@article{tucs2023quantum,
  title={Quantum annealing designs nonhemolytic antimicrobial peptides in a discrete latent space},
  author={Tucs, Andrejs and Berenger, Francois and Yumoto, Akiko and Tamura, Ryo and Uzawa, Takanori and Tsuda, Koji},
  journal={ACS Med. Chem. Lett.},
  volume={14},
  number={5},
  pages={577--582},
  year={2023},
  publisher={ACS Publications}
}

@article{inoue2022towards,
  title={Towards optimization of photonic-crystal surface-emitting lasers via quantum annealing},
  author={Inoue, Takuya and Seki, Yuya and Tanaka, Shu and Togawa, Nozomu and Ishizaki, Kenji and Noda, Susumu},
  journal={Opt. Express},
  volume={30},
  number={24},
  pages={43503--43512},
  year={2022},
  publisher={Optica Publishing Group}
}

@article{mohseni2022ising,
  title={Ising machines as hardware solvers of combinatorial optimization problems},
  author={Mohseni, Naeimeh and McMahon, Peter L and Byrnes, Tim},
  journal={Nat. Rev. Phys.},
  volume={4},
  number={6},
  pages={363--379},
  year={2022}
}

@misc{FixAE,
  author = {{Fixstars Amplify}},
  title = {{Fixstars Amplify Annealing Engine}},
  note = {\url{https://amplify.fixstars.com/en/}}
}

@article{kirkpatrick1983optimization,
  title={Optimization by simulated annealing},
  author={Kirkpatrick, Scott and Gelatt, C Daniel and Vecchi, Mario P},
  journal={Science},
  volume={220},
  number={4598},
  pages={671--680},
  year={1983},
  publisher={American association for the advancement of science}
}

@article{johnson1991optimization,
  title={Optimization by simulated annealing: {A}n experimental evaluation; part {II}, graph coloring and number partitioning},
  author={Johnson, David S and Aragon, Cecilia R and McGeoch, Lyle A and Schevon, Catherine},
  journal={Oper. Res.},
  volume={39},
  number={3},
  pages={378--406},
  year={1991},
  publisher={INFORMS}
}

@book{das2005quantum,
  title={Quantum annealing and related optimization methods},
  author={Das, Arnab and Chakrabarti, Bikas K},
  volume={679},
  year={2005},
  publisher={Springer Science \& Business Media}
}

@article{goto2019combinatorial,
  title={Combinatorial optimization by simulating adiabatic bifurcations in nonlinear {H}amiltonian systems},
  author={Goto, Hayato and Tatsumura, Kosuke and Dixon, Alexander R},
  journal={Sci. Adv.},
  volume={5},
  number={4},
  note={{A}rt. no. eaav2372},
  year={2019},
  publisher={American Association for the Advancement of Science}
}

@inproceedings{okuyama2017ising,
  title={An {I}sing computer based on simulated quantum annealing by path integral {M}onte {C}arlo method},
  author={Okuyama, Takuya and Hayashi, Masato and Yamaoka, Masanao},
  booktitle={2017 IEEE international conference on rebooting computing (ICRC)},
  pages={1--6},
  year={2017},
  organization={IEEE}
}

@article{tsukamoto2017accelerator,
  title={An accelerator architecture for combinatorial optimization problems},
  author={Tsukamoto, Sanroku and Takatsu, Motomu and Matsubara, Satoshi and Tamura, Hirotaka},
  journal={Fujitsu Scientific and Technical Journal},
  volume={53},
  number={5},
  pages={8--13},
  year={2017}
}

@article{goto2021high,
  title={High-performance combinatorial optimization based on classical mechanics},
  author={Goto, Hayato and Endo, Kotaro and Suzuki, Masaru and Sakai, Yoshisato and Kanao, Taro and Hamakawa, Yohei and Hidaka, Ryo and Yamasaki, Masaya and Tatsumura, Kosuke},
  journal={Sci. Adv.},
  volume={7},
  number={6},
  note={{Art.} no. eabe7953},
  year={2021},
  publisher={American Association for the Advancement of Science}
}

@article{Johnson2011,
  author={Johnson, Mark W and Amin, Mohammad HS and Gildert, Suzanne and Lanting, Trevor and Hamze, Firas and Dickson, Neil and Harris, Richard and Berkley, Andrew J and Johansson, Jan and Bunyk, Paul and Chapple, E M and Enderud, C and Hilton, J P and Karimi, K and Ladizinsky, E and Ladizinsky, N and Oh, T and Perminov, I and Rich, C and Thom, M C and Tolkacheva, E and Truncik, C J S and Uchaikin, S and Wang, J and Wilson, B and Rose, G},
  issn     = {00280836},
  journal  = {Nature},
  number   = {7346},
  pages    = {194--198},
  title    = {{Quantum annealing with manufactured spins}},
  volume   = {473},
  year     = {2011}
}

@article{kadowaki1998quantum,
  title={Quantum annealing in the transverse {I}sing model},
  author={Kadowaki, Tadashi and Nishimori, Hidetoshi},
  journal={Phys. Rev. E},
  volume = {58},
  issue = {5},
  pages = {5355--5363},
  year={1998},
  publisher={APS}
}

@article{farhi2000quantum,
  title={Quantum computation by adiabatic evolution},
  author={Farhi, Edward and Goldstone, Jeffrey and Gutmann, Sam and Sipser, Michael},
  journal={arXiv preprint quant-ph/0001106},
  year={2000}
}

@article{dirks2004paradigms,
  title={Paradigms for computational nucleic acid design},
  author={Dirks, Robert M and Lin, Milo and Winfree, Erik and Pierce, Niles A},
  journal={Nucleic Acids Res.},
  volume={32},
  number={4},
  pages={1392--1403},
  year={2004},
  publisher={Oxford University Press}
}

@article{ward2023fitness,
  title={Fitness functions for {RNA} structure design},
  author={Ward, Max and Courtney, Eliot and Rivas, Elena},
  journal={Nucleic Acids Res.},
  volume={51},
  number={7},
  pages={e40--e40},
  year={2023},
  publisher={Oxford University Press}
}

@article{hofacker1994fast,
  title={Fast folding and comparison of {RNA} secondary structures},
  author={Hofacker, Ivo L and Fontana, Walter and Stadler, Peter F and Bonhoeffer, L Sebastian and Tacker, Manfred and Schuster, Peter},
  journal={Monats. Chem.},
  volume={125},
  pages={167--188},
  year={1994},
  publisher={SPRINGER VERLAG}
}

@article{lorenz2011viennarna,
  title={Vienna{RNA} Package 2.0},
  author={Lorenz, Ronny and Bernhart, Stephan H and H{\"o}ner zu Siederdissen, Christian and Tafer, Hakim and Flamm, Christoph and Stadler, Peter F and Hofacker, Ivo L},
  journal={Algorithms Mol. Biol.},
  volume={6},
  pages={1--14},
  year={2011},
  publisher={Springer}
}

@article{morris2014rise,
  title={The rise of regulatory {RNA}},
  author={Morris, Kevin V and Mattick, John S},
  journal={Nat. Rev. Genet,},
  volume={15},
  number={6},
  pages={423--437},
  year={2014},
  publisher={Nature Publishing Group UK London}
}

@article{crick1970central,
  title={Central dogma of molecular biology},
  author={Crick, Francis},
  journal={Nature},
  volume={227},
  number={5258},
  pages={561--563},
  year={1970},
  publisher={Nature Publishing Group UK London}
}

@article{doudna2002chemical,
  title={The chemical repertoire of natural ribozymes},
  author={Doudna, Jennifer A and Cech, Thomas R},
  journal={Nature},
  volume={418},
  number={6894},
  pages={222--228},
  year={2002},
  publisher={Nature Publishing Group UK London}
}

@article{serganov2007ribozymes,
  title={Ribozymes, riboswitches and beyond: regulation of gene expression without proteins},
  author={Serganov, Alexander and Patel, Dinshaw J},
  journal={Nat. Rev, Genet.},
  volume={8},
  number={10},
  pages={776--790},
  year={2007},
  publisher={Nature Publishing Group UK London}
}

@article{reese2005oligo,
  title={Oligo-and poly-nucleotides: 50 years of chemical synthesis},
  author={Reese, Colin B},
  journal={Org. Biom. Chem.},
  volume={3},
  number={21},
  pages={3851--3868},
  year={2005},
  publisher={Royal Society of Chemistry}
}

@article{pardi2018mrna,
  title={{mRNA} vaccines—a new era in vaccinology},
  author={Pardi, Norbert and Hogan, Michael J and Porter, Frederick W and Weissman, Drew},
  journal={Nat. Rev. Drug Discov.},
  volume={17},
  number={4},
  pages={261--279},
  year={2018},
  publisher={Nature Publishing Group}
}

@article{hamada2018silico,
  title={In silico approaches to {RNA} aptamer design},
  author={Hamada, Michiaki},
  journal={Biochimie},
  volume={145},
  pages={8--14},
  year={2018},
  publisher={Elsevier}
}

@article{singh2017exploring,
  title={Exploring the potential of genome editing {CRISPR-Cas9} technology},
  author={Singh, Vijai and Braddick, Darren and Dhar, Pawan Kumar},
  journal={Gene},
  volume={599},
  pages={1--18},
  year={2017},
  publisher={Elsevier}
}

@article{jaffrey2018rna,
  title={{RNA}-based fluorescent biosensors for detecting metabolites in vitro and in living cells},
  author={Jaffrey, Samie R},
  journal={Adv. Pharmacol.},
  volume={82},
  pages={187--203},
  year={2018},
  publisher={Elsevier}
}

@article{bauer2006engineered,
  title={Engineered riboswitches as novel tools in molecular biology},
  author={Bauer, Gesine and Suess, Beatrix},
  journal={J. Biotechnol.},
  volume={124},
  number={1},
  pages={4--11},
  year={2006},
  publisher={Elsevier}
}

@article{dixon2010reengineering,
  title={Reengineering orthogonally selective riboswitches},
  author={Dixon, Neil and Duncan, John N and Geerlings, Torsten and Dunstan, Mark S and McCarthy, John EG and Leys, David and Micklefield, Jason},
  journal={Proc. Natl. Acad. Sci.},
  volume={107},
  number={7},
  pages={2830--2835},
  year={2010},
  publisher={National Academy of Sciences}
}

@article{seeman1976rna,
  title={{RNA} double-helical fragments at atomic resolution: {I}. {T}he crystal and molecular structure of sodium adenylyl-3', 5'-uridine hexahydrate},
  author={Seeman, Nadrian C and Rosenberg, John M and Suddath, FL and Kim, Jung Ja Park and Rich, Alexander},
  journal={J. Mol. Biol.},
  volume={104},
  number={1},
  pages={109--144},
  year={1976},
  publisher={Elsevier}
}

@article{rosenberg1976rna,
  title={{RNA} double-helical fragments at atomic resolution: {II}. The crystal structure of sodium guanylyl-3', 5'-cytidine nonahydrate},
  author={Rosenberg, John M and Seeman, Nadrian C and Day, Roberta O and Rich, Alexander},
  journal={J. Mol. Biol.},
  volume={104},
  number={1},
  pages={145--167},
  year={1976},
  publisher={Elsevier}
}

@article{varani2000g,
  title={The {G}{\textperiodcentered} {U} wobble base pair},
  author={Varani, Gabriele and McClain, William H},
  journal={EMBO Rep.},
  volume={1},
  pages={18--23},
  year={2000},
  publisher={John Wiley \& Sons, LtdChichester, UK}
}

@article{andronescu2004new,
  title={A new algorithm for {RNA} secondary structure design},
  author={Andronescu, Mirela and Fejes, Anthony P and Hutter, Frank and Hoos, Holger H and Condon, Anne},
  journal={J. Mol. Biol.},
  volume={336},
  number={3},
  pages={607--624},
  year={2004},
  publisher={Elsevier}
}

@article{zadeh2011nucleic,
  title={Nucleic acid sequence design via efficient ensemble defect optimization},
  author={Zadeh, Joseph N and Wolfe, Brian R and Pierce, Niles A},
  journal={J. Comput. Chem.},
  volume={32},
  number={3},
  pages={439--452},
  year={2011},
  publisher={Wiley Online Library}
}

@article{zadeh2011nupack,
  title={{NUPACK}: Analysis and design of nucleic acid systems},
  author={Zadeh, Joseph N and Steenberg, Conrad D and Bois, Justin S and Wolfe, Brian R and Pierce, Marshall B and Khan, Asif R and Dirks, Robert M and Pierce, Niles A},
  journal={J. Comput. Chem.},
  volume={32},
  number={1},
  pages={170--173},
  year={2011},
  publisher={Wiley Subscription Services, Inc., A Wiley Company Hoboken}
}

@article{busch2006info,
  title={{INFO-RNA}—a fast approach to inverse {RNA} folding},
  author={Busch, Anke and Backofen, Rolf},
  journal={Bioinformatics},
  volume={22},
  number={15},
  pages={1823--1831},
  year={2006},
  publisher={Oxford University Press}
}

@article{taneda2010modena,
  title={{MODENA}: a multi-objective {RNA} inverse folding},
  author={Taneda, Akito},
  journal={Adv. Appl. Bioinform. Chem.},
  pages={1--12},
  year={2010},
  publisher={Taylor \& Francis}
}

@article{lyngso2012frnakenstein,
  title={Frnakenstein: multiple target inverse {RNA} folding},
  author={Lyngs{\o}, Rune B and Anderson, James WJ and Sizikova, Elena and Badugu, Amarendra and Hyland, Tomas and Hein, Jotun},
  journal={BMC Bioinformatics},
  volume={13},
  number={1},
  pages={260},
  year={2012},
  publisher={Springer}
}

@article{rubio2018multiobjective,
  title={Multiobjective metaheuristic to design {RNA} sequences},
  author={Rubio-Largo, {\'A}lvaro and Vanneschi, Leonardo and Castelli, Mauro and Vega-Rodr{\'\i}guez, Miguel A},
  journal={IEEE Trans. Evol. Comput.},
  volume={23},
  number={1},
  pages={156--169},
  year={2019},
  publisher={IEEE}
}

@article{garcia2013rnaifold,
  title={{RNAiFOLD}: a constraint programming algorithm for {RNA} inverse folding and molecular design},
  author={Garcia-Martin, Juan Antonio and Clote, Peter and Dotu, Ivan},
  journal={J. Bioinform. Comput. Biol.},
  volume={11},
  number={02},
  pages={1350001},
  year={2013},
  publisher={World Scientific}
}

@article{yang2017rna,
  title={{RNA} inverse folding using {Monte Carlo} tree search},
  author={Yang, Xiufeng and Yoshizoe, Kazuki and Taneda, Akito and Tsuda, Koji},
  journal={BMC Bioinformatics},
  volume={18},
  number={1},
  pages={468},
  year={2017},
  publisher={Springer}
}

@article{kleinkauf2015antarna,
  title={{antaRNA}: ant colony-based {RNA} sequence design},
  author={Kleinkauf, Robert and Mann, Martin and Backofen, Rolf},
  journal={Bioinformatics},
  volume={31},
  number={19},
  pages={3114--3121},
  year={2015},
  publisher={Oxford University Press}
}

@article{eastman2018solving,
  title={Solving the {RNA} design problem with reinforcement learning},
  author={Eastman, Peter and Shi, Jade and Ramsundar, Bharath and Pande, Vijay S},
  journal={PLOS Comput. Biol.},
  volume={14},
  number={6},
  pages={e1006176},
  year={2018},
  publisher={Public Library of Science San Francisco, CA USA}
}

@article{runge2018learning,
  title={Learning to design {RNA}},
  author={Runge, Frederic and Stoll, Danny and Falkner, Stefan and Hutter, Frank},
  journal={arXiv preprint arXiv:1812.11951},
  year={2018}
}

@article{shi2018sentrna,
  title={{SentRNA}: Improving computational {RNA} design by incorporating a prior of human design strategies},
  author={Shi, Jade and Das, Rhiju and Pande, Vijay S},
  journal={arXiv preprint arXiv:1803.03146},
  year={2018}
}

@article{koodli2019eternabrain,
  title={{EternaBrain}: automated {RNA} design through move sets and strategies from an internet-scale {RNA} videogame},
  author={Koodli, Rohan V and Keep, Benjamin and Coppess, Katherine R and Portela, Fernando and Eterna participants and Das, Rhiju},
  journal={PLOS Comput. Biol.},
  volume={15},
  number={6},
  pages={e1007059},
  year={2019},
  publisher={Public Library of Science San Francisco, CA USA}
}

@article{sumi2024deep,
  title={Deep generative design of {RNA} family sequences},
  author={Sumi, Shunsuke and Hamada, Michiaki and Saito, Hirohide},
  journal={Nat. Methods},
  volume={21},
  number={3},
  pages={435--443},
  year={2024},
  publisher={Nature Publishing Group US New York}
}

@article{akiyama2022informative,
  title={Informative {RNA} base embedding for {RNA} structural alignment and clustering by deep representation learning},
  author={Akiyama, Manato and Sakakibara, Yasubumi},
  journal={NAR Genom. Bioinform.},
  volume={4},
  number={1},
  pages={lqac012},
  year={2022},
  publisher={Oxford University Press}
}

@article{jones1998efficient,
  title={Efficient global optimization of expensive black-box functions},
  author={Jones, Donald R and Schonlau, Matthias and Welch, William J},
  journal={J. Glob. Optim.},
  volume={13},
  number={4},
  pages={455--492},
  year={1998},
  publisher={Springer}
}

@article{forrester2009recent,
  title={Recent advances in surrogate-based optimization},
  author={Forrester, Alexander IJ and Keane, Andy J},
  journal={Prog. Aerosp. Sci.},
  volume={45},
  number={1-3},
  pages={50--79},
  year={2009},
  publisher={Elsevier}
}

@article{ramu2022survey,
  title={A survey of machine learning techniques in structural and multidisciplinary optimization},
  author={Ramu, Palaniappan and Thananjayan, Pugazhenthi and Acar, Erdem and Bayrak, Gamze and Park, Jeong Woo and Lee, Ikjin},
  journal={Struct. Multidiscip. Optim.},
  volume={65},
  number={9},
  pages={266},
  year={2022},
  publisher={Springer}
}

@article{tamura2024machine,
  title={Machine learning prediction of the mechanical properties of injection-molded polypropylene through {X}-ray diffraction analysis},
  author={Tamura, Ryo and Nagata, Kenji and Sodeyama, Keitaro and Nakamura, Kensaku and Tokuhira, Toshiki and Shibata, Satoshi and Hammura, Kazuki and Sugisawa, Hiroki and Kawamura, Masaya and Tsurimoto, Teruki and others},
  journal={Sci. Technol. Adv. Mater.},
  volume={25},
  number={1},
  pages={2388016},
  year={2024},
  publisher={Taylor \& Francis}
}

@article{nawa2023quantum,
  title={Quantum annealing optimization method for the design of barrier materials in magnetic tunnel junctions},
  author={Nawa, Kenji and Suzuki, Tsuyoshi and Masuda, Keisuke and Tanaka, Shu and Miura, Yoshio},
  journal={Phys. Rev. Appl.},
  volume={20},
  issue={2},
  note={{Art.} no. 024044},
  year={2023},
  publisher={APS}
}

@article{kim2024quantum,
  title={Quantum annealing-aided design of an ultrathin-metamaterial optical diode},
  author={Kim, Seongmin and Park, Su-Jin and Moon, Seunghyun and Zhang, Qiushi and Hwang, Sanghyo and Kim, Sun-Kyung and Luo, Tengfei and Lee, Eungkyu},
  journal={Nano Converg.},
  volume={11},
  number={1},
  pages={16},
  year={2024},
  publisher={Springer}
}

@article{couzinie2025machine,
  title={Machine learning supported annealing for prediction of grand canonical crystal structures},
  author={Couzini{\'e}, Yannick and Seki, Yuya and Nishiya, Yusuke and Nishi, Hirofumi and Kosugi, Taichi and Tanaka, Shu and Matsushita, Yu-ichiro},
  journal={J. Phys. Soc. Jpn.},
  volume={94},
  number={4},
  pages={044802},
  year={2025},
  publisher={The Physical Society of Japan}
}

@article{tamura2021performance,
  title={Performance comparison of typical binary-integer encodings in an {I}sing machine},
  author={Tamura, Kensuke and Shirai, Tatsuhiko and Katsura, Hosho and Tanaka, Shu and Togawa, Nozomu},
  journal={IEEE Access},
  volume={9},
  pages={81032--81039},
  year={2021},
  publisher={IEEE}
}

@article{chen2021performance,
  title={Performance of domain-wall encoding for quantum annealing},
  author={Chen, Jie and Stollenwerk, Tobias and Chancellor, Nicholas},
  journal={IEEE Trans. Quantum Eng.},
  volume={2},
  pages={1--14},
  year={2021},
  publisher={IEEE}
}

@article{kikuchi2024performance,
  title={Performance of Domain-Wall Encoding in Digital {I}sing Machine},
  author={Kikuchi, Shuta and Takahashi, Kotaro and Tanaka, Shu},
  journal={arXiv preprint arXiv:2410.11198},
  year={2024}
}

@article{chancellor2019domain,
  title={Domain wall encoding of discrete variables for quantum annealing and {QAOA}},
  author={Chancellor, Nicholas},
  journal={Quantum Sci. Technol.},
  volume={4},
  number={4},
  note={{Art.} no. 045004},
  year={2019},
  publisher={IOP Publishing}
}

@article{anderson2016principles,
  title={Principles for predicting {RNA} secondary structure design difficulty},
  author={Anderson-Lee, Jeff and Fisker, Eli and Kosaraju, Vineet and Wu, Michelle and Kong, Justin and Lee, Jeehyung and Lee, Minjae and Zada, Mathew and Treuille, Adrien and Das, Rhiju and others},
  journal={J. Mol. Biol.},
  volume={428},
  number={5},
  pages={748--757},
  year={2016},
  publisher={Elsevier}
}

@article{huang2024tutorial,
  title={Tutorial: {AI}-assisted exploration and active design of polymers with high intrinsic thermal conductivity},
  author={Huang, Xiang and Ju, Shenghong},
  journal={J. Appl. Phys.},
  volume={135},
  number={17},
  year={2024},
  publisher={AIP Publishing}
}

@article{blank2020pymoo,
  title={pymoo: Multi-Objective Optimization in {Python}},
  author={Blank, Julian and Deb, Kalyanmoy},
  journal={IEEE Access},
  volume={8},
  pages={89497--89509},
  year={2020},
  publisher={IEEE}
}

@article{mathews2004incorporating,
  title={Incorporating chemical modification constraints into a dynamic programming algorithm for prediction of {RNA} secondary structure},
  author={Mathews, David H and Disney, Matthew D and Childs, Jessica L and Schroeder, Susan J and Zuker, Michael and Turner, Douglas H},
  journal={Proc. Natl. Acad. Sci.},
  volume={101},
  number={19},
  pages={7287--7292},
  year={2004},
  publisher={National Academy of Sciences}
}

@article{turner2010nndb,
  title={{NNDB}: the nearest neighbor parameter database for predicting stability of nucleic acid secondary structure},
  author={Turner, Douglas H and Mathews, David H},
  journal={Nucleic Acids Res.},
  volume={38},
  number={suppl\_1},
  pages={D280--D282},
  year={2010},
  publisher={Oxford University Press}
}

@inproceedings{schnall2008inverting,
  title={Inverting the {Viterbi} algorithm: an abstract framework for structure design},
  author={Schnall-Levin, Michael and Chindelevitch, Leonid and Berger, Bonnie},
  booktitle={Proceedings of the 25th international conference on machine learning},
  pages={904--911},
  year={2008}
}

@article{bonnet2020designing,
  title={Designing {RNA} secondary structures is hard},
  author={Bonnet, {\'E}douard and Rz{\k{a}}{\.z}ewski, Pawe{\l} and Sikora, Florian},
  journal={J. Comput. Biol.},
  volume={27},
  number={3},
  pages={302--316},
  year={2020},
  publisher={Mary Ann Liebert, Inc., publishers 140 Huguenot Street, 3rd Floor New~…}
}

@article{koshikawa2025efficient,
  title={Efficient bit labeling in factorization machines with annealing for traveling salesman problem},
  author={Koshikawa, Shota and Hosaka, Aruto and Yoshida, Tsuyoshi},
  journal={Sci. Rep.},
  volume={15},
  number={1},
  pages={26910},
  year={2025},
  publisher={Nature Publishing Group UK London}
}

\section*{Acknowledgements}
This work was partially supported by the Japan Society for the Promotion of Science (JSPS) KAKENHI (Grant Number JP23H05447), the Council for Science, Technology, and Innovation (CSTI) through the Cross-ministerial Strategic Innovation Promotion Program (SIP), ``Promoting the application of advanced quantum technology platforms to social issues'' (Funding agency: QST), Japan Science and Technology Agency (JST) (Grant Number JPMJPF2221).
S. Tanaka wishes to express their gratitude to the World Premier International Research Center Initiative (WPI), MEXT, Japan, for their support of the Human Biology-Microbiome-Quantum Research Center (Bio2Q).
The computations in this work were partially performed using the facilities of the Supercomputer Center, the Institute for Solid State Physics, The University of Tokyo.

\section*{Author information}
\subsection*{Contributions}
S.K. conducts conceptualization, data curation, formal analysis, investigation, methodology, software, visualization, writing--original draft, and writing--review \& editing.
S.T. conducts conceptualization, funding acquisition, project administration, resources, supervision, and writing--review \& editing.
All authors reviewed the manuscript. 

\section*{Ethics declarations}
\subsection*{Competing interests}
The authors declare no competing interests.

\section*{Additional information}
\subsection*{Publisher’s note}
Springer Nature remains neutral with regard to jurisdictional claims in published maps and institutional affiliations.

\end{document}